\documentclass[journal,twoside,web]{IEEEtran}
 \usepackage{balance}   
\usepackage{mathtools}
\usepackage{amsmath,amssymb,bbm}
\usepackage{amsthm}     

\usepackage{enumitem}
\usepackage{cancel}
\usepackage{graphicx}
\usepackage{subcaption}
\usepackage{pifont}
\usepackage{xspace}
\usepackage{xcolor}

\usepackage{makecell}
\usepackage{mathrsfs}
\usepackage{mathtools}
\usepackage{cite}
\newtheorem{assumption}{Assumption}
\newtheorem{lemma}{Lemma}

\newtheorem{remark}{Remark}

\newtheorem{theorem}{Theorem}

\newcommand{\E}{\mathbb{E}}

\newcommand{\mb}{\mathrm{mb}}

\newcommand{\cO}{\mathcal{O}}

\newcommand{\R}{\mathbb{R}}
\newcommand{\ind}{\mathbf{1}}
\newcommand{\norm}[1]{\left\| #1 \right\|}

\allowdisplaybreaks 

\usepackage{algorithm}
\usepackage{algpseudocode}
\algrenewcommand\algorithmicrequire{\textbf{Input:}}
\algrenewcommand\algorithmicensure{\textbf{Output:}}

\newcommand{\CommentState}[1]{\Statex\hspace{\algorithmicindent}{\color{blue}// #1}}

\makeatletter\@mparswitchfalse\makeatother\normalmarginpar 
\usepackage[textwidth=1cm, textsize=scriptsize]{todonotes}

\def\BibTeX{{\rm B\kern-.05em{\sc i\kern-.025em b}\kern-.08em
    T\kern-.1667em\lower.7ex\hbox{E}\kern-.125emX}}
\title{A Communication-Efficient Decentralized Actor-Critic Algorithm}

\author{%
	Xiaoxing~Ren$^{1}$, Nicola~Bastianello$^{2}$,  Thomas~Parisini$^{3,4,5}$, Andreas A. Malikopoulos~$^{1}$
	\thanks{The work of X.R. and A.M. was supported in part by NSF under Grants CNS-2401007, CMMI-2348381, IIS-2415478, and in part by MathWorks.}
\thanks{The work of N.B. was partially supported by the European Union’s Horizon Research and Innovation Actions programme under grant agreement No. 101070162.}
	\thanks{$^{1}$ Systems Engineering, Cornell Robotics, Applied Math, and School of Civil and Environmental
Engineering, Cornell University, Ithaca, NY,
USA. (email: xr49@cornell.edu, amaliko@cornell.edu)}%
	\thanks{$^{2}$School of Electrical Engineering and Computer Science, and Digital Futures, KTH Royal Institute of Technology, Stockholm, Sweden. (email: nicolba@kth.se)}
 	\thanks{$^{3}$Department of Electrical and Electronic Engineering, Imperial College London, London, United Kingdom. (email: t.parisini@imperial.ac.uk)}
	\thanks{$^{4}$Department of Electronic Systems, Aalborg University, Denmark.}%
	\thanks{$^{5}$Department of Engineering and Architecture, University of Trieste, Trieste, Italy.}%
}

\begin{document}

\maketitle

\begin{abstract}
In this paper, we study the problem of reinforcement learning in multi-agent systems where communication among agents is limited. We develop a decentralized actor–critic learning framework in which each agent performs several local updates of its policy and value function, where the latter is approximated by a multi-layer neural network, before exchanging information with its neighbors. 
This local training strategy substantially reduces the communication burden while maintaining coordination across the network. We establish finite-time convergence analysis for the algorithm under Markov-sampling.
 Specifically, to attain the $\varepsilon$-accurate stationary point, the sample complexity is of order $\mathcal{O}(\varepsilon^{-3})$ and the communication complexity is of order $\mathcal{O}(\varepsilon^{-1}\tau^{-1})$, where $\tau$ denotes the number of local training steps. We also show how the final error bound depends on the neural network's approximation quality.
Numerical experiments in a cooperative control setting illustrate and validate the theoretical findings.

\end{abstract}


\section{Introduction}
\label{sec:introduction}
Multi-agent reinforcement learning (MARL) has emerged as a powerful framework for solving complex decision-making problems in dynamic environments. It has been successfully applied to multi-player games~\cite{mnih2015human,silver2016mastering}, autonomous driving~\cite{Sumanth2021,sun2025recommendation}, powertrain systems~\cite{Malikopoulos2010a,Malikopoulos2011}, intelligent systems~\cite{zhang2021intelligent}, wireless networks~\cite{zhao2019deep}, and multi-robot coordination~\cite{duan2016benchmarking}. 
A particularly important subclass is \emph{cooperative MARL}, where multiple decentralized agents interact with a shared environment, each observing local information and receiving local rewards. The collective objective is to learn a set of local policies that maximize a global performance measure.

Another closely related line of research is \emph{multi-task reinforcement learning}~\cite{zeng2021decentralized,yu2020gradient}, where the goal is to learn a unified policy that generalizes across multiple tasks rather than optimizing for a single objective. Recent developments have also sought to integrate learning and control to handle such multi-task and multi-agent settings~\cite{Malikopoulos2022a,Malikopoulos2024}.

\subsection{Background on Reinforcement Learning}

Classical single-agent reinforcement learning methods can be broadly categorized into \emph{value-based} and \emph{policy gradient} approaches. 
Value-based methods, such as temporal-difference and $Q$-learning, iteratively update value estimates using observed rewards to infer an optimal policy~\cite{barto2021reinforcement,bertsekas2019reinforcement}. 
In contrast, policy gradient methods directly optimize a parameterized policy by ascending the gradient of the expected return in a model-free manner~\cite{sutton1999policy}. 
Prominent examples include REINFORCE~\cite{williams1992simple} and GPOMDP~\cite{baxter2001infinite}. 

Natural policy gradient (NPG) methods~\cite{kakade2001natural,bagnell2003covariant} further improve stability by incorporating the intrinsic geometry of the policy space through the Kullback--Leibler (KL) divergence. 
Trust region policy optimization~\cite{schulman2015trust} combines NPG with a line search procedure to guarantee improvement, and proximal policy optimization~\cite{schulman2017proximal} replaces the KL constraint with a simplified objective using a penalty term or clipping.

Although effective, policy gradient methods often suffer from high variance and slow convergence because Monte Carlo rollouts are used to estimate the value function required for computing the policy gradient
The \emph{actor--critic} framework addresses this issue by learning a baseline value function (the critic) to reduce gradient variance while the actor updates the policy parameters~\cite{konda1999actor}. 
This structure improves stability and sample efficiency and has been widely adopted in both single- and multi-agent settings~\cite{haarnoja2018soft,peters2008natural,bhatnagar2007incremental,konda1999actor_type}. 

\subsection{Decentralized Actor--Critic Learning}

In fully decentralized MARL, agents communicate only with neighboring nodes over a network and do not rely on a central coordinator. 
Each agent observes its own reward, which may differ across the network. 
Several decentralized actor--critic algorithms have been proposed~\cite{zeng2022learning,lin2019communication,zhang2018fully,mao2025decentralized,hairi2022finite,suttle2020multi,dai2025distributed,li2022finite,chen2022sample}. 
The work in~\cite{zhang2018fully} first established asymptotic convergence under a two-timescale update scheme with diminishing step sizes. 
Subsequent efforts improved \emph{communication efficiency}, for example, by reducing exchanged variables to a few scalar quantities~\cite{lin2019communication}. 
More recent studies have investigated \emph{finite-time analysis} under constant step sizes~\cite{hairi2022finite,mao2025decentralized,li2022finite,chen2022sample}, while~\cite{dai2025distributed} incorporated neural approximators for distributed value estimation and proved global convergence. 

Despite these advances, \emph{provably efficient decentralized actor--critic algorithms} that achieve both communication and sample efficiency under nonlinear function approximation remain limited. 

\subsection{Overview of This Work}

In this paper, we develop a communication-efficient decentralized actor--critic algorithm that utilizes local training.  We provide a finite-time analysis of its sample and communication complexities under Markovian sampling. 

Our main contributions are as follows:
\begin{itemize}
    \item \textit{Communication efficiency:} We build on a local-training alternating direction method of multipliers (ADMM) framework that requires only a single round of communication of local policy parameters after several local update steps. Only policy parameters are shared, preserving privacy.   
    \item  \textit{Mini-batch updates:} We adopt Markovian mini-batch sampling for both actor and critic updates, reducing variance and improving sample efficiency compared with single-sample updates. 
    \item  \textit{Neural function approximation:} We employ multi-layer neural networks for the critic, extending theoretical guarantees beyond linear approximations. 
    \item \textit{Finite-time analysis:} 
    We establish finite-time convergence guarantees under fixed step sizes, quantifying the trade-off between communication frequency and accuracy, and showing that the final error bound depends on the neural network’s approximation quality. 
    \item \textit{Empirical validation:} Experiments on a cooperative navigation task demonstrate the algorithm’s superior communication efficiency relative to existing methods.  
\end{itemize}

We have provided a broad overview of existing MARL methods in this section. 
The following subsection offers a more focused comparison with closely related works, emphasizing the differences in algorithmic structures, convergence guarantees, and complexity analyses.

\subsection{Comparison with the State of the Art}
\label{subsec:sota}

Multi-agent reinforcement learning algorithms face inherent challenges associated with high communication costs. 
To address this issue, several studies have explored communication-efficient strategies that reduce the communication burden. 
One class of approaches limits communication by transmitting only partial parameter information at each iteration, which substantially decreases the communication cost~\cite{lin2019communication}. 
Another line of work employs quantization techniques to compress exchanged messages and thereby reduce bandwidth consumption~\cite{chen2024communication}. 
Alternative approach improves efficiency by replacing single-sample updates with mini-batch updates, leading to lower complexities and reduced variance~\cite{chen2022sample}.  
The decentralized single-loop actor--critic algorithm~\cite{li2022finite} permits agents to communicate only once every $K_c$ iterations during the critic update while maintaining convergence guarantees, where $K_c$ can be any integer within a given interval. 
However, it still requires frequent reward exchanges during the actor update, which can impose substantial communication overhead.  

Recent developments in distributed optimization and learning have further improved communication efficiency through \emph{local training}, in which agents perform several local updates before synchronizing with their peers or a coordinator~\cite{hien_nguyen_performance_2023,liu_decentralized_2023,alghunaim_local_2023,guo_randcom_2023,ren2025communication}. 
Among these approaches, the local-training ADMM framework~\cite{ren2025communication} combines distributed ADMM and local training to achieve a favorable communication--computation trade-off. 
The present work extends this framework to the decentralized MARL setting, where temporal dependencies, stochastic state transitions, and policy optimization introduce significant analytical challenges absent in static distributed optimization.  

Many existing MARL algorithms employ single-sample estimators for both actor and critic updates, which often results in high-variance gradient estimates and slow convergence~\cite{zhang2018fully,lin2019communication,dai2025distributed}. 
To enhance sample efficiency, more recent methods adopt online actor--critic schemes based on Markovian mini-batch updates, where samples are drawn from a single trajectory~\cite{chen2022sample,hairi2022finite,mao2025decentralized,li2022finite,xu2020improving}. 
The proposed algorithm also follows this design principle, using Markovian mini-batches to improve the convergence behavior.  

From a theoretical perspective, most convergence analyses of decentralized actor--critic algorithms remain confined to \emph{linear} value function approximations~\cite{zhang2018fully,lin2019communication,chen2022sample,hairi2022finite,mao2025decentralized,li2022finite,zeng2022learning}, which restricts their applicability to complex or nonlinear environments. 
Recent studies have begun extending theoretical guarantees to neural function approximators in the \emph{single-agent} setting~\cite{tian2023performance,tian2023convergence_convergence}, while distributed formulations with nonlinear critics remain largely unexplored. 
An exception is the work in~\cite{dai2025distributed}, which establishes global convergence for a distributed neural policy gradient method employing a single-layer network to approximate the $Q$-function. 
In contrast, the present work provides finite-time convergence guarantees for decentralized actor--critic algorithms that utilize multi-layer fully connected neural networks for value function approximation, thereby extending theoretical guarantees to more expressive function classes.

\begin{table*}[!t]
\centering
\caption{Comparison with the state of the art related decentralized actor-critic algorithms.}
\label{tab:comparison}
    \begin{tabular}{cccccccc}
    \hline
    Algorithm [Ref.] & \thead{Sampling \\ scheme} & \thead{Function \\ approximation} & \thead{Sample \\ complexity} & \thead{Shared\\informatio}  & \thead{Communication \\ complexity} & \thead{convergence} \\
    \hline
    
   \cite{lin2019communication} &  \makecell{Markovian \\ single sample} & Linear & \text{-} & \makecell{Local critic  parameter \\Local TD-error} & \text{-} & asymptotic \\
   
    \cite{chen2022sample} & \makecell{Markovian \\ mini-batch} & Linear & $\mathcal{O}(\varepsilon^{-2}\ln \varepsilon^{-1})$ & \thead{Local critic parameter\\ noisy local reward} &  $\mathcal{O}(\varepsilon^{-1}\ln \varepsilon^{-1})$ & Finite-time \\

    \cite{hairi2022finite} & \makecell{Markovian \\ mini-batch} & Linear & $\mathcal{O}\!\left(\varepsilon^{-2}\ln \varepsilon^{-1}\right)$ & \makecell{Local critic  parameter \\Local TD-error} &  $\mathcal{O}(\varepsilon^{-1}\ln \varepsilon^{-1})$ & Finite-time \\

    \cite{li2022finite} & \makecell{Markovian \\ sampling} & Linear & $\tilde{\mathcal{O}}(\varepsilon^{-2})$ &\thead{Local critic parameter\\ noisy local reward} & $\tilde{\mathcal{O}}(\varepsilon^{-2} \ln \varepsilon^{-1})$ & Finite-time \\

   \cite{mao2025decentralized} & \makecell{Markovian \\ mini-batch} & Linear &  $\mathcal{O}(\varepsilon^{-2}\ln \varepsilon^{-1})$ &\thead{Local critic parameter\\ noisy local reward} &   $\mathcal{O}(\varepsilon^{-1}\ln \varepsilon^{-1})$ & Finite-time \\

    \cite{dai2025distributed} & \makecell{i.i.d$^\dagger$ \\ Markovian mini-batch}  & \makecell{Single-layer \\ neural network} & \text{-} & Local critic parameter & \text{-} & Finite-time \\

    \hline

 [this work] & \makecell{Markovian \\ mini-batch}  & \makecell{Multi-layer \\ neural network} &  $\mathcal{O}(\varepsilon^{-3} )$  & Local policy parameter  & $\mathcal{O}(\varepsilon^{-1}\tau^{-1})$ & Finite-time \\
    \hline
    \end{tabular}
    \\\vspace{0.1cm}
    \text{-} not given in the paper \\
    $^\dagger$ i.i.d. sample in distributed critic update, Markov sample in decentralized actor update
\end{table*}

Table~\ref{tab:comparison} summarizes the comparison between our algorithm and state-of-the-art methods. While most prior studies employ linear or single-layer function approximation, our framework provides theoretical guarantees under multi-layer neural networks. The communication complexity bound explicitly includes the number of local training steps $\tau$, enabling reduced communication rounds for a given accuracy $\varepsilon<1$ by selecting $\tau > \frac{1}{\ln{\varepsilon^{-1}}}$.  This reveals a clear trade-off between local computation and communication. Furthermore, our algorithm ensures finite-time convergence, extending beyond the asymptotic guarantees of previous works.
Finally, our method enjoys finite-time convergence guarantees, in contrast to earlier asymptotic results. Collectively, these features highlight the strengthened theoretical guarantees and improved efficiency of our framework.

\subsubsection*{Notation} We denote by $\nabla f$ the gradient of a differentiable function $f$. 
For a matrix $A \in \mathbb{R}^{n \times n}$, let $\lambda_{\min}(A)$ and $\lambda_{\max}(A)$ denote its smallest and largest eigenvalues, respectively. 
Given $n \in \mathbb{N}$, let $\mathbf{1}_n \in \mathbb{R}^n$ be the vector with all elements equal to one, $\mathbf{I} \in \mathbb{R}^{n \times n}$ the identity matrix, and $\mathbf{0} \in \mathbb{R}^{n \times n}$ the zero matrix. 
The standard inner product of two vectors $x, y \in \mathbb{R}^n$ is written as $\langle x, y \rangle = \sum_{h=1}^n x_h y_h$, and $\| \cdot \|$ denotes the Euclidean norm for vectors and the matrix-induced $2$-norm for matrices. 
Throughout the paper, the notation $\tilde{\mathcal{O}}$ is used to suppress polylogarithmic factors.

\section{Problem Formulation}\label{sec:problem-formulation}

We consider a team of $N$ agents, indexed by $\mathscr{V} = \{1, \dots, N\}$, that interact with each other and with a stochastic environment over a discrete time horizon $t = 0, 1, \dots, T$. 
The communication topology among the agents is represented by an undirected graph $\mathscr{G} = (\mathscr{V}, \mathcal{E})$, where $\mathcal{E} = \{(i,j) \mid i, j \in \mathscr{V} \ \text{and} \ (i,j) \ \text{is an edge}\}$. 
Each agent observes the global state--action pair and receives a local reward that is not shared with other agents. 
The following subsection introduces the main components and notation underlying the MARL problem considered in this work.

\subsubsection{State and action} 
    We denote by $\boldsymbol{S}^i$ and $\boldsymbol{A}^i$ the finite local state and finite action spaces of agent $i \in \mathscr{V}$, respectively. Then $s^i \in \boldsymbol{S}^i$ and $a^i \in \boldsymbol{A}^i$ denote the local state and local action of agent $i$, respectively.
Collecting all the local states and actions, we denote the global state and action as 
\begin{align*}
    s &= (s^1, \ldots, s^N) \in \boldsymbol{S} = \boldsymbol{S}^1 \times \ldots \times \boldsymbol{S}^N \in \R^{d_s} \\
    a &= (a^1, \ldots, a^N) \in \boldsymbol{A} = \boldsymbol{A}^1 \times \ldots \times \boldsymbol{A}^N \in \R^{d_a}.
\end{align*}
Throughout the paper we will denote by $s_t = (s^1_{t}, \ldots, s^N_{t})$ and $a_t = (a^1_{t}, \ldots, a^N_{t})$ as the global state and action at time $t$.

\subsubsection{State transition probability} 
We model the system dynamics in terms of a \emph{Markov transition kernel} 
$\mathcal{P}(\cdot \mid s, a): \mathcal{B}(\boldsymbol{S}) \to [0,1]$, 
where $\mathcal{B}(\boldsymbol{S})$ denotes the Borel $\sigma$-algebra on the state space $\boldsymbol{S}$. 
For any measurable set $B \subseteq \boldsymbol{S}$, the transition kernel is defined as
\begin{equation}
    \mathcal{P}(B \mid s, a) = 
    \mathbb{P}\big(s_{t+1} \in B \mid s_t = s, a_t = a\big).
    \label{eq:state-transition-prob}
\end{equation}
When $\boldsymbol{S}$ is finite, the kernel reduces to a \emph{state transition probability matrix} 
$\mathcal{P}(s' \mid s, a): \boldsymbol{S} \times \boldsymbol{A} \times \boldsymbol{S} \to [0,1]$, 
representing the probability of moving from state $s$ to state $s'$ under action $a$.



\subsubsection{Observations and policy}
At each time step $t$, agent $i \in \mathscr{V}$ obtains a local observation denoted by $o_t^i$. 
The agent selects actions according to a stochastic policy $\pi_{\omega^i}^i(a^i \mid o_t^i)$, which maps local observations to probability distributions over actions, where $\omega^i \in \mathbb{R}^{d_i}$ denotes the parameter vector of agent $i$'s policy~\cite{Malikopoulos2021}. 

Throughout this work, we consider the case in which the global state $s_t$ is fully observable by all agents. 
Under this assumption, each observation coincides with the global state, i.e., $o_t^i = s_t$ for all $i \in \mathscr{V}$. 
The resulting joint policy over all agents then factorizes as
\begin{equation}
    \pi_{\omega}(a \mid s) = \prod_{i=1}^{N} \pi_{\omega^i}^i(a^i \mid s), \label{eq:joint-policy}
\end{equation}
where $\omega = (\omega^1, \dots, \omega^N) \in \mathbb{R}^{d}$, with $d = \sum_{i=1}^{N} d_i$ denoting the dimension of the joint policy parameter vector.

\subsubsection{Reward function}
We denote by $r^i(s, a): \boldsymbol{S} \times \boldsymbol{A} \rightarrow \R$ the local reward function for agent $i$ after action $a$ is taken at state $s$. For simplicity we also denote by $r^i_{t} = r^i(s_t, a_t)$
the instantaneous reward for agent $i$ at time $t$, and $\bar{r}_t = \frac{1}{N} \sum_{i=1}^N r^i_{t} $
as the average reward across all agents at time $t$.

The discounted value function of the policy $\pi$ is defined as:
\begin{equation} \label{eq:true_value}
\mathcal{V}_{\pi}(s) = \mathbb{E}_{s,a \sim \pi} \left[ \sum_{t=0}^{\infty} \gamma^{t} r(s_t, a_t) \right],
\end{equation}
where $\mathbb{E}_{s,a \sim \pi}$ stands for the expectation when the starting state is $s$ and actions are taken according to policy $\pi$, $0 < \gamma < 1$ is a discount factor.

\subsection{System Model and Policy Structure}
In this paper, the global state is assumed to be common knowledge and is observed by all agents, whereas the local reward functions remain private to individual agents. The agents’ objective is, therefore, to collaboratively learn a policy that maximizes the global long-term return. This objective is formally captured in the following problem formulation.

We denote by $J(\omega)$ the discounted average cumulative reward under the joint policy $\pi_{{\omega}}$, which is given by
\begin{equation}
J(\omega) = \mathbb{E}_{s \sim s_{\text{ini}}} \left[ \sum_{t=0}^{H-1} \gamma^t \bar{r}_t \;\middle|\; s_0 = s,\; a_t \sim \pi_{{\omega}}(\cdot \mid s_t) \right],
\end{equation}
where $s_{\text{ini}}$ is the distribution of the initial state $s_0$, $\bar{r}_t = \frac{1}{N} \sum_{i=1}^N r^i_{t} $, and $H$ is the trajectory horizon.

Thus, the objective is to find the optimal policy that maximizes the average cumulative reward, i.e., to solve:
\begin{equation} \label{main_problem}
\max_{{{\omega}} \in \R^d} J({{\omega}}) = \frac{1}{N}\sum_{i=1}^N J_i({{\omega}}).
\end{equation}
where
$J_i({{\omega}}) = \mathbb{E}_{s \sim  s_{\text{ini}}} \left[ \sum_{t=0}^{H-1} \gamma^t {r}^i_t \;\middle|\; s_0 = s,\; a_t \sim \pi_{{\omega}}(\cdot \mid s_t) \right]$.

\subsection{Decentralized optimization setup}
Let each agent $i$ hold a copy $
{\omega}^i$ of the  global policy parameter ${\omega}$, when the network $\mathscr{G}$ is connected. Hence, problem~\eqref{main_problem} can be rewritten as follows:
\begin{equation} \label{decen_problem}
\max_{{\omega}^1, \dots, {\omega}^N \in \R^d} \frac{1}{N} \sum_{i=1}^n J_i({\omega}^i) 
\quad \text{s.t} \quad {\omega}^i = {\omega}^j, \ \text{for all } (i, j) \in \mathcal{E}.
\end{equation}
The consensus constraints guarantee the agents reach the same solution as the centralized problem \eqref{main_problem}.

\begin{assumption}\label{as:graph}
$\mathit{\mathscr{G}} = (\mathscr{V},\mathcal{E})$ is a connected, undirected graph.
\end{assumption}
\begin{assumption} \label{ass:r_bounded}
There exists $R_{\max} > 0$ such that for any agent $i \in \mathscr{V}$ and any Markovian sample $(s,a) \in \boldsymbol{S} \times \boldsymbol{A} $, we have
\[
0 \leq r^{i}(s,a) \leq R_{\max}.
\]
\end{assumption}


The following assumption ensures that the state will get closer and closer to the stationary distribution as \( t \) increases.
Let \( \mathcal{P}^t \) denote the matrix \(\mathcal{P} \) raised to the \( t \)'th power, \( \mathcal{P}^t_{s;\cdot} \) be the row of \( \mathcal{P}^t \) corresponding to state \( s \), and \( \|\cdot\|_{\mathrm{TV}} \) is the total variation distance between probability measures.
\begin{assumption}\label{ass:ergodicity}
The Markov chain transition probability matrix $\mathcal{P}$ induced by policy $\pi$ satisfies:
\begin{align}
\sup_{s \in \boldsymbol{S}} \left\| \mathcal{P}^t_{s; \cdot} - \mu \right\|_{\mathrm{TV}}  \leq \kappa  \zeta^t, \quad \forall t \geq 0,
\end{align}
with $\kappa > 0$ and $\zeta \in (0,1)$, $\mu$ is the stationary distribution. Additionally, we assume $\mu$ is strictly positive, i.e., there exists $\mu_{l}>0$ such that $\mu \ge \mu_{l}$.
\end{assumption}
Assumption~\ref{ass:ergodicity} is standard in the literature on reinforcement learning and temporal-difference methods~\cite{xu2020improved,ma2020variance,chen2022sample,li2022finite}. 
As established in~\cite{levin2017markov}, this assumption holds when the underlying Markov chain is irreducible and aperiodic, and it guarantees geometric convergence of the state distribution to its stationary distribution.


\section{Algorithm Design}




\subsection{Nonlinear approximation of the value function}
In actor-critic methods, the critic is an estimator of the value function, which is used to calculate the policy gradient. Since the true value function~\eqref{eq:true_value} is, of course, not accessible to the agents, we aim to design a suitable approximation of it. 

Without loss of generality, we employ a multi-layer neural network to serve as a functional approximation of $\mathcal{V}_{\pi}$.
Specifically, we use a network with $L_v$ layers characterized by
\begin{align*}
    x^{\ell} = \frac{1}{\sqrt{m}} \varrho \left( \theta^{\ell} x^{\ell-1} \right),  \ell \in \{1, 2, \ldots, L_v\}, \quad \text{and} \ x^0 \in \boldsymbol{S} \, ,
\end{align*}
where $m \in \mathbb{N}$ is the common width of all layers, $\varrho$ is an activation function, and the weights are denoted as $\theta^0 \in \R^{m\times d_s}$, $\theta^{\ell} \in \R^{m \times m}, \ell =1,2,\ldots, L_v -1$.
Hence, the nonlinear value function approximation of~\eqref{eq:true_value} is given by
\begin{equation} \label{eq:nonlinear_value_function}
V(s, \theta) = \frac{1}{\sqrt{m}} b^\top x^{L_v} \, ,
\end{equation}
where $b \in \R^{m}$.

The neural network is thus characterized by the values of $\theta$, the vector collecting all the 
 \( \theta^{1}, \ldots, \theta^{L_v} \) of the different layers, and $b$.
We assume that each entry of $\theta$ and $b$ is initialized from $\mathcal{N}(0 , 1)$,  $b$ is kept fixed during training. Therefore, the agents are only tasked with training the weights $\theta$; in the following, its norm is defined by $\|\theta\|^2 = \sum_{\ell=1}^{L_v} \|\theta^{\ell}\|_F^2$,
where \( \|\cdot\|_F \) is the Frobenius norm.
This definition of the neural network, together with the setting of a fixed weight vector $b$, has been widely adopted in the literature, including works on TD learning \cite{tian2023performance,xu2020finite} and deep neural networks \cite{liu2020linearity}.

\smallskip

We introduce several assumptions that will be used in section~\ref{sec:convergence} to analyze the convergence of the critic update in the algorithm proposed in section~\ref{subsec:algorithm}.
%

\begin{assumption}
\label{ass:s_bounded}
All states are uniformly bounded; that is,
\[
\| s \| \le 1, \quad \forall\, s \in \boldsymbol{S}.
\]
\end{assumption}

\begin{assumption} \label{ass:sigma_smooth}
The activation function \( \varrho \) is \( L_{\varrho} \)-Lipschitz and \( c_{\varrho} \)-smooth.
\end{assumption}
\begin{assumption} \label{ass:x_bounded}
For all \( \ell \in \{1, 2, \ldots, L_v\} \), 
$|x^{\ell}_h| $ is $\tilde{\mathcal{O}}(1)$  at initialization. Here, \( x^{\ell}_h \) means the \( h\)-th entry of \( x^{\ell} \).
\end{assumption}
\begin{assumption} \label{ass:initial_bounded}
The initialization of the network parameters is not too large, that is, for all \( h \in \{1, 2, \ldots, L_v\} \), $\| \theta_0^{h} \| \leq \mathcal{O}(\sqrt{m})$.
\end{assumption}

Let us denote the log-likelihood policy gradient as
\begin{equation}
\psi_{\mathbf{\omega} }(a | s) = \nabla_{{\omega}}\ln \pi_{\omega}(a | s).
\end{equation}
We now introduce the following assumption used in \cite{xu2020improved} for the Lipschitz continuity of policy gradients.
\begin{assumption} \label{assu:smoothness}
There exist constants $C_\psi, C_H > 0$ such that for all $\omega \in \R^d $, $s \in \boldsymbol{S}$ and $a \in \boldsymbol{A}$, the gradient and Hessian of the log-density of the policy function satisfy
\begin{equation}
\begin{aligned}
&\| \psi_\omega(a|s) \| \leq C_\psi, \quad \| \nabla^2_{\mathbf{\omega} }\ln \pi_{\omega}(a | s) \| \leq C_H.
\end{aligned}
\end{equation}
\end{assumption}

Define the compact set $\Theta= \mathcal{B}(\theta_{0,0}, \tilde{\theta}) = \{\theta \mid \|\theta - \theta_{0,0}\| \leq \tilde{\theta}\}$ onto which we project our critic parameter, where $\theta_{0,0}$ is the initial critic parameter.
We use a parametric approximator \(V:\boldsymbol{S} \times\Theta\to\R\).
For any fixed \(\theta\in\Theta\), we write
\[
V_\theta:\boldsymbol{S} \to\mathbb{R},\qquad
V_\theta(s)\;=\;V(s,\theta).
\]
That is, \(V_\theta\) denotes the function of \(s\) obtained by plugging \(\theta\) into the approximator.
Given the policy class $\Pi$ and bounded parameter set $\Theta$,
define
\begin{equation} \label{eq:value_approx}
\begin{aligned}
\varsigma_{\mathrm{approx}} &= \sup_{\pi\in\Pi}\ \inf_{\theta\in\Theta}\ \|V_\theta-\mathcal{V}_\pi\|_\infty
\\&= \sup_{\pi\in\Pi}\ \inf_{\theta\in\Theta}\ \max_{s\in\boldsymbol{S}}
|V(\theta, s)-\mathcal{V}_\pi(s)|.
\end{aligned}
\end{equation}
For each $\pi$, let 
$\hat\theta_\pi \in \arg\min_{\theta\in\Theta}\|V_\theta-\mathcal{V}_\pi\|_\infty$.
Then, for all $\pi\in\Pi$, $V_{\hat{\theta}_{\pi}}$ is a $\varsigma_{\mathrm{approx}}$-approximation of the true value function $\mathcal{V}_{\pi}$ such that
\begin{equation} \label{eq:value_approximation_accuracy}
    \max_{s \in \boldsymbol{S}}  | V(s, \hat{\theta}_\pi) - \mathcal{V}_{\pi}(s) | \leq \varsigma_{\mathrm{approx}}.
\end{equation}

\begin{remark}
Assumptions~\ref{ass:s_bounded}, \ref{ass:sigma_smooth}, \ref{ass:x_bounded}, and \ref{ass:initial_bounded} 
are commonly employed to characterize the properties of neural network function approximators~\cite{tian2023performance,tian2023convergence_convergence}. 
As discussed in~\cite{tian2023performance}, Assumptions~\ref{ass:s_bounded}, \ref{ass:x_bounded}, and \ref{ass:initial_bounded} 
can typically be satisfied through appropriate scaling or network initialization. 
Assumption~\ref{ass:sigma_smooth} requires smooth activation functions and is therefore fulfilled by commonly used activations such as \textit{sigmoid}, \textit{arctan}, \textit{tanh}, and \textit{ELU}. 
Furthermore, Assumption~\ref{assu:smoothness} is standard in the literature to ensure the Lipschitz continuity of policy gradients~\cite{xu2020improved,chen2024decentralized,yang2021sample,fatkhullin2023stochastic}.
\end{remark}

\begin{remark}
The parameter set $\Theta$ is defined as a closed ball centered at the initial parameter, with radius $\tilde{\theta}$ selected through empirical tuning. 
In implementation, $\tilde{\theta}$ is chosen sufficiently large to enable effective learning while avoiding numerical instability. 
The term $\varsigma_{\mathrm{approx}}$ denotes the inherent approximation error of the neural network. 
Although it is not directly observable or controllable, $\varsigma_{\mathrm{approx}}$ depends on the representational capacity of the chosen function approximator and serves as a theoretical quantity for bounding the algorithm’s performance.
\end{remark}

\subsection{Proposed algorithm}\label{subsec:algorithm}
Here, we propose a novel communication- and sample-efficient decentralized actor-critic algorithm to solve the problem formulated in Section~\ref{sec:problem-formulation}.

Our decentralized actor–critic method is built upon the LT-ADMM \cite{ren2025communication}.
The key features we leverage are: (i) \emph{local training}: each agent approximately
solves its subproblem by running $\tau$ gradient steps on local data before communication;
(ii) \emph{bridge variables} $\{z_{ij,k}\}$ on edge $(i,j)$ to enforce consensus on the policy parameters;
and (iii) a single \emph{communication round} per outer iteration $k$, controlled by the penalty parameter $\rho$.
Given a connected graph, at round $k$, agent $i$ updates its local policy parameter
$\omega^i$ by approximately solving the ADMM proximal subproblem
\begin{equation}\label{eq:ltadmm-subprob}
\omega^i_{k+1} \approx \arg\min_{\omega^i \in \R^d}\;\Big\{ - J_i(\omega^i) 
+ \frac{\rho}{2}\sum_{j\in\mathcal{N}_i}\big\|\omega_i - z_{i j, k} \big\|^2 \Big\},
\end{equation}
where $\mathcal{N}_i $ denotes the neighbors of agent $i$.
After local training, bridge variables are updated on edges:
\begin{equation}\label{eq:bridge-update}
z_{i j, k+1} = \frac{1}{2} \left(z_{i j, k} - z_{j i, k}+2 \rho \omega^{j}_{k+1}\right)
    \end{equation}
This is the standard LT-ADMM pattern with one communication per iteration.
The local training updates in MARL are as follows:
\begin{itemize}
    \item {Critic (value) update:}
Each agent $i$ uses a neural network defined in \eqref{eq:nonlinear_value_function} to approximate the value function and updates $\theta^i$ via projected TD learning \cite{sutton1988learning} with Markovian mini-batches of size $N_c$:
\begin{align}
\delta^i_{t} &= r^i_{t} + \gamma V(s_{t+1}, \theta^i) - V(s_t, \theta^i), \nonumber\\
\theta^i_{t+1} \;\;&\leftarrow\; \theta^i_{t} + \eta \cdot \frac{1}{N_c}\sum_{t}\delta^i_{t}\,\nabla_{\theta} V(s_t, \theta^i), \label{eq:critic-update}
\end{align}
where $N_c$ is the Markovian mini-batch collected during round $k$ and $\eta>0$ is the critic step size.
The TD error $\delta^i_{t}$ will also serve as a low-variance advantage estimate in the actor step.
\item  {Actor (policy) update:}
Conditioned on $\theta^i_t$ from~\eqref{eq:critic-update}, agent $i$ forms a stochastic policy-gradient estimator with Markovian mini-batches of size $B$:
\begin{equation}\label{eq:pg-estimator}
g_i(\omega^i_t) \;=\; \frac{1}{B}\sum_{t} \psi_{t}^{i}(a_{t} | s_{t}) \widehat{A}^{i}_{t},
\quad \widehat{A}^{i}_{t}=\delta^{i}_{t},
\end{equation}
\item Agent $i$ performs $\tau$ local gradient steps to approximately solve~\eqref{eq:ltadmm-subprob}
with initialization $\omega^i_{k,0}=\omega^i_{k}$ and fixed actor stepsizes.
We set $\omega^i_{k+1}=\omega^i_{k,\tau}$ and then apply the edge update~\eqref{eq:bridge-update}. This realizes the LT-ADMM ``compute locally, communicate once" principle in the MARL setting. Following~[34], the Markovian mini-batch $B$ improves both sample efficiency and estimator accuracy.
\end{itemize}

We summarize the full procedure in Algorithm~1. The main hyperparameters are the number of local updates $\tau$, the actor learning rates $\alpha$ and $\beta$, the critic learning rate $\eta$, the consensus penalty parameter $\rho$, the mini-batch size $B$ and $N_c$, and the critic update iterations $T_c$. As noted in \cite{tian2023performance}, the stationary distribution $\mu_{\phi^i_{k,t}}$ used in the local critic update can be approximated by generating a sufficiently long path from $\mathcal{P}$ and ignoring the influence of initial states.

\begin{algorithm} 
\caption{LT-ADMM Actor-Critic}
\label{alg: LT-ADMM Actor-Critic}
\begin{algorithmic}[1]
\Require Number of agents $N$, training episodes $K$ and critic update numbers $ T_c$, local training number $\tau$,  batch size $B$ and $N_c$, learning rates $\alpha$, $\beta$ and $\eta$, penalty parameter $\rho$.  Initial policy parameter $\mathbf{\omega}_0$, critic parameter $\theta_{0,0}$, the critic projection ball $\Theta= \mathcal{B}(\theta_{0,0}, \omega) = \{\theta \mid \|\theta - \theta_{0,0}\| \leq \tilde{\theta}\}$.
\State \textbf{Initialization:} For each agent $i$, set $z_{ij, 0}= z_{ji, 0}=\omega^i_0$, 
\For{$k = 0, 1, \ldots, K-1$}
\CommentState{Local Training}

set $\phi_{k,0}^i = {\omega}^{i}_{k}$
\For{$t =0, \ldots, \tau-1$}

\For{$i = 1,\ldots,N$} 
\CommentState{local critic update}

$s_{k, tB}, {\theta}_{k,t} =$ \text{Decentralized TD}($s_{ini}$, $ \phi_{k,t}$, $N_c$, $\eta$, $T_c$, $\Theta$), $s^i_{ini}$ is sampled from  the  stationary distribution $\mu_{\phi^i_{k,t}}$
\State 
\CommentState{collect $B$ Markovian samples} 
  \For{$q = 0, \dots, B-1$}
    
$a_{k, tB+q} \sim \pi_{\phi^i_{k,t}}(\cdot \mid s_{k, tB+q}), \quad  s_{k, tB+q+1} \sim \mathcal{P}_{\pi_{\phi^i_{k,t}}}(\cdot \mid s_{k, tB+q}, a_{k, tB+q})$, observe reward $r^i_{k, tB+q}$
    \EndFor
     \CommentState{local actor update}
\begin{equation}
\begin{aligned}
&g_i(\phi_{k,t}^i)   = \frac{1}{B} \sum_{l=tB}^{(t+1)B-1}
\\& \left[
{r}_{k,l}^{i} + \gamma V(s_{k,l+1}, \theta_{k,t}^{i}  )
-  V(s_{k,l}, \theta_{k,t}^{i}  )
\right] \psi_{k,t}^{i}(a_{k,l} | s_{k,l}),
\end{aligned}
\end{equation}
\begin{equation}
\phi_{k,t+1}^{i}  =  \phi^i_{k,t}+ \alpha g_i(\phi_{k,t}^i) -  \beta (   (\rho\left|\mathcal{N}_i\right| \omega_{k}^i-\sum_{j \in \mathcal{N}_i} z_{i j, k} ) )
\end{equation}
\EndFor
\EndFor

Set
${\omega}^{i}_{k+1}  =\phi_{k,\tau}^i$
\CommentState{communication and auxiliary update}
 \State transmit $z_{ij,k} - 2 \rho \omega_{i,k+1}$ to each neighbor $j \in \mathcal{N}_i$, and receive the corresponding transmissions
\[
z_{i j, k+1} = \frac{1}{2} \left(z_{i j, k} - z_{j i, k}+2 \rho \omega^{j}_{k+1}\right)
    \]
\EndFor
\State \textbf{Output:} $\boldsymbol{\omega}^{i}_{K}$, $i =1,\ldots,N$.
\end{algorithmic}
\end{algorithm}
\begin{algorithm}
    \caption{Decentralized TD ($s_{ini}$, $\phi$, $N_c$, $\eta$, $T_c$, $\Theta$)}
\label{alg:decentralized_td}
\textbf{Initialize:} Critic parameter ${\theta}_{k,t,0}= {\theta}_{k,t-1} $

\begin{algorithmic}[1]
\For{critic iterations $t' = 0, 1, \ldots, T_c - 1$}
    \For{agents $i = 1, \ldots, N$ \textbf{in parallel}}
    
       $s_{t',0} = s_{t'-1,N_c}$ if $t'\geq 1$ else $s_{t',0} = s^i_{ini}$

      \CommentState {collect $N_c$ Markovian samples}
    
    \For{$j = 0, \dots, N_c-1$}
    
    $a_{t',j} \sim \pi_{\phi^i}(\cdot \mid s_{t',j}), \quad 
    s_{t',j+1} \sim \mathcal{P}_{\pi_{\phi^i}}(\cdot \mid s_{t',j}, a_{t',j})$ \;
    (observe rewards $r^i_{t',j}$) \;
    \EndFor
\CommentState{decentralized TD update}
\State \begin{equation}
\begin{aligned}
&\theta_{k,t,t'+1}^{i} = \theta_{k,t,t'}^i
+ \frac{\eta}{N_c} \sum_{j=0}^{N_c - 1} 
[r_{t',j}^{i} + \gamma  V(s_{t',j+1}, \theta_{k,t,t'}^{i} )  \\& - V(s_{t',j}, \theta_{k,t,t'}^{i} )] \nabla_{\theta} V(s_{t',j}, \theta_{k,t,t'}^{i} ),
\\& \theta_{k,t,t'+1}^{i} = \textbf{Proj}_{\Theta} (\theta_{k,t,t'+1}^{i})
\end{aligned}
\end{equation}
%
    \EndFor
\EndFor
\State Uniformly pick index $d' \in [1,T_c]$, set $\theta_{k,t} = \theta_{k,t,d' }$.
\State \textbf{Output:}  $s_{T_c-1, N_{c}}$, $\theta_{k,t} $
\end{algorithmic}
\end{algorithm}

\section{Convergence Analysis and Discussion}\label{sec:convergence}

The following result characterizes the convergence of the proposed Algorithm~\ref{alg: LT-ADMM Actor-Critic}.

\begin{theorem} \label{theorem:main}
Let Assumptions~\ref{as:graph},   \ref{ass:r_bounded}, \ref{ass:ergodicity},  \ref{ass:s_bounded}, \ref{ass:sigma_smooth}, \ref{ass:x_bounded}, \ref{ass:initial_bounded}, and \ref{assu:smoothness} hold. Choose learning rates that satisfy $\alpha \leq \bar{\alpha}$ and $\frac{1}{\tau\lambda_u\rho} \leq \beta <  \frac{2}{\tau\lambda_u\rho}$, where $\bar{\alpha}$ is defined in \eqref{gamma_sgd} in Appendix~\ref{sec:preliminary_definition} and $\lambda_u$ is the largest eigenvalue of the Laplacian matrix of $\mathit{\mathscr{G}}$. Then the following holds:
   \begin{equation} \label{eq:converge}
\begin{aligned}
 \frac{1}{K} \sum_{k=0}^{K-1} \mathcal{D}_k &\leq \frac{16 ( {J}\left(\bar{\omega}_K\right) - {J}\left(\bar{\omega}_0\right)   )}{\alpha \tau K} +  \mathcal{O}\left(\frac{\|\hat{\mathbf{d}}_0\|^2}{ NK\lambda_l\rho^2} \right) 
\\& + \mathcal{O}( \frac{\epsilon_{\theta}^2}{\eta T_c} + \frac{1}{B}  )+ \mathcal{O}(\varsigma_{\mathrm{approx}} + \varsigma_{\mathrm{approx}}^2 +\frac{1}{\sqrt{m}}) 
\\&+ \mathcal{O}(\eta \varsigma_{\mathrm{approx}}^2) +  \mathcal{O}(\eta),
\end{aligned}
\end{equation}
where $\bar{\omega}_k = \frac{1}{N} \sum_{i = 1}^N {\omega}^i_k$,  $\lambda_l$ is the smallest nonzero eigenvalue of the Laplacian matrix of $\mathit{\mathscr{G}}$, $\hat{\mathbf{d}}_0$ is related to the initial condition as defined in Lemma~\ref{lem:devitaion_aver}, $\epsilon_{\theta} = 2\tilde{\theta}$ 
and the convergence metric $\mathcal{D}_k$ is defined as
\begin{equation}\label{eq:convergence-metric}
\mathcal{D}_k = \mathbb{E} \left[ \left\|\nabla J\left(\bar{\omega}_{k}\right)\right\|^2+\frac{1}{\tau} \sum_{t = 0}^{\tau-1} \|{\frac{1}{N} \sum_{i = 1}^N g_i\left(\phi_{k,t}^i\right)}\|^2 \right].
\end{equation}
\end{theorem}
\begin{proof}
The proof is detailed in the appendix. 
\end{proof}

\smallskip

\subsection{Discussion}
\paragraph{Convergence metric}
In Theorem~\ref{theorem:main}, we adopt $\mathcal{D}_k$ as the convergence metric. This quantity characterizes the optimality gap of the nonconvex problem~\eqref{decen_problem} and becomes zero when the algorithm reaches a stationary point. Furthermore, as shown in~\eqref{sum_d} in Appendix~\ref{appen:main_result}, a bounded $\mathcal{D}_k $ also implies a bounded consensus error.

\paragraph{Convergence rate and complexity}
Inspecting the bound in Theorem~\ref{theorem:main}, we see that the convergence rate depends on the network connectivity and the mini-batch size $B$. In particular, better network connectivity (larger algebraic connectivity $\lambda_l$) and larger mini-batch size $B$ lead to faster convergence.
Besides, as seen in \eqref{gamma_sgd}, $\bar{\alpha}$ is proportional to $\lambda_l$. Thus, better connected graphs (larger $\lambda_l$) also result in larger learning rate bounds.

Moreover, in order to achieve $\mathbb{E}\!\left[\|\mathcal{D}_K \|^2\right]\le \varepsilon$
for any $\varepsilon > \mathcal{O}(\varsigma_{\mathrm{approx}} + \varsigma_{\mathrm{approx}}^2 +\frac{1}{\sqrt{m}})$, we can choose $K = \mathcal{O}(\tau^{-1}\varepsilon^{-1})$, $B=\mathcal{O}(\varepsilon^{-1})$, $\eta = \mathcal{O}(\varepsilon)$
and $T_c=\mathcal{O}(\varepsilon^{-2})$.
Consequently, the overall sample complexity is 
$K\tau(T_cN_c+B)=\mathcal{O}(\varepsilon^{-3 })$,
while the communication complexity for synchronizing local policy parameters is $K=\mathcal{O}(\tau^{-1}\varepsilon^{-1})$.

\paragraph{Asymptotic error}
Theorem~\ref{theorem:main} provides the finite-time analysis of the decentralized actor-critic algorithm under Markovian sampling.
With the critic learning rate chosen as $\eta = \frac{1}{\sqrt{T_c}}$, and sufficiently large $T_c$ and $B$, the final error is bounded by $\mathcal{O} ( \varsigma_{\mathrm{approx}} + \varsigma_{\mathrm{approx}}^2 + \frac{1}{\sqrt{m}} )$.
   \paragraph{Comparison with \cite{ren2025communication}} 
   A key distinction between our algorithm and LT-ADMM~\cite{ren2025communication} lies in the nature of the gradient estimators. In actor–critic methods, the policy gradient estimator is inherently biased because the critic depends on approximate value function estimates rather than exact returns. Consequently, unlike in~\cite{ren2025communication}, where an unbiased stochastic gradient enables variance $\sigma$ to be controlled through step-size adjustment, the bias in our setting prevents direct reduction of $\sigma$ (defined in \eqref{eq:sigma}) via tuning the actor learning rate. Furthermore, unlike LT-ADMM, which assumes i.i.d. data, reinforcement learning operates with temporally correlated Markovian samples, necessitating additional mixing conditions to ensure convergence. Finally, while distributed learning typically focuses on empirical risk minimization, multi-agent reinforcement learning (MARL) aims to optimize the expected long-term return through policy updates, introducing non-stationarity as the sampling distribution evolves with the policy. These fundamental differences make both the algorithm design and the theoretical analysis considerably more challenging than in the LT-ADMM framework.

\section{Experiments}
We evaluate our method on the \emph{Cooperative Navigation} task adopted in MARL studies \cite{qu2019value,jiang2022mdpgt,chen2024decentralized}.
The environment contains $N$ agents and $N$ landmarks placed in a bounded $2\times 2$ square. 
Every agent is randomly initialized and assigned a unique target landmark. 
Agents aim to reach their own targets while avoiding collisions with other agents.

The global state $s$ consists of the $2D$ positions of all agents and their landmarks; the action space is discrete,
$\boldsymbol{A}^i=\{\text{up},\text{down},\text{left},\text{right},\text{stay}\}$.
Let $p^i_t$ denote the position of agent $i$ at time $t$ and $\ell_i$ its assigned landmark.
For each movable agent $i$ with mass $m_i$ and damping coefficient ${\kappa}_{\text{damp}}\in[0,1)$, let $F_t^{i}\in\mathbb{R}^2$ be the applied force corresponding to the action $a_t^i$ at time $t$. 
Velocities are first damped, then accelerated by the force, then clipped to a maximum speed $v_{\max}^{i}$, and finally, positions are updated:
\begin{align}
\tilde v_{t+1}^{i} &= (1-{\kappa}_{\text{damp}})\, v_t^{i} \;+\; \frac{\Delta t}{m_i}\, F_t^{i}, \label{eq:dyn-preclip}\\
v_{t+1}^{i} &= 
\begin{cases}
\tilde v_{t+1}^{i}, & \|\tilde v_{t+1}^{i}\| \le v_{\max}^{i},\\[2pt]
\dfrac{v_{\max}^{i}}{\|\tilde v_{t+1}^{i}\|}\, \tilde v_{t+1}^{i}, & \text{otherwise,}
\end{cases} \label{eq:dyn-clip}\\
p_{t+1}^{i} &= p_t^{i} + \Delta t\, v_{t+1}^{i}. \label{eq:dyn-pos-update}
\end{align}
Agents that choose the action 'stay' keep their states unchanged.
The agent $i$'s reward is
\begin{equation} \label{eq:coopnav-reward}
r_t^{i}
= - \bigl\|p_t^{i}-\ell_i\bigr\|
  - \sum_{j\neq i} \ind\!\left\{\|p_t^{i}-p_t^{j}\| < d_{\mathrm{coll}}\right\},
\end{equation}
i.e., the negative distance to its landmark with an additional unit penalty upon collision. All positions lie in $[-1,1]$ along each dimension. The task is considered done when the sum of all agent–landmark distances is less than $0.15$.
The team's objective is to maximize the total return of all agents, i.e., $\sum_{i=1}^5 r_t^i$. 
The experiments were conducted on a desktop computer equipped with an Intel Core~i9-13900KF CPU (3.0~GHz, 24~cores), 64~GB RAM, running Windows~11~Pro (64-bit). The implementation was based on Python and PyTorch.

We compare with state-of-the-art decentralized actor-critic methods~\cite{hairi2022finite,chen2022sample,li2022finite} and the decentralized policy-gradient method that also exchanges only policy parameters, multi-agent decentralized NPG (MDNPG)~\cite{chen2024decentralized}. 
All experiments use a ring graph with $N=5$.  For Algorithm~\ref{alg: LT-ADMM Actor-Critic}, we set $B=N_c=20$, $\tau = 3$, $T_c = 3$, $\rho = 0.5$, $\alpha = 0.001$, $\beta = 0.01$, and $\eta = 0.001$.
To be fair, for all algorithms, we employ a policy network and a state value network, each with two hidden layers and ReLU as the activation function, with a Markov mini-batch sample size of $20$ and mini-batch size of $3$ in the critic update. The training episodes for all algorithms are $20000$.

We account for the communication complexity of each algorithm when scaling the $x$-axis, in particular, the total communication rounds. 
In each episode, the algorithms in \cite{hairi2022finite,chen2022sample,li2022finite} employ multiple averaging steps for critic updates (e.g., for critic parameters, noisy local rewards, or TD errors). In our comparative simulations, we set this number to $2$ for \cite{chen2022sample} and $3$ for \cite{hairi2022finite,li2022finite}. Similarly, \cite{chen2024decentralized} uses two averaging steps for actor updates (one for gradient tracking and one for parameter update). In contrast, Algorithm~\ref{alg: LT-ADMM Actor-Critic} requires only one communication per episode for actor updates. We refer to Algorithm 2 in \cite{hairi2022finite} as TD-sharing-AC, Algorithm 2 in \cite{li2022finite} as Single-timescale, and Algorithm 1 in \cite{chen2022sample} as SC-efficient.

\begin{figure}
    \centering
    \includegraphics[scale=0.28]{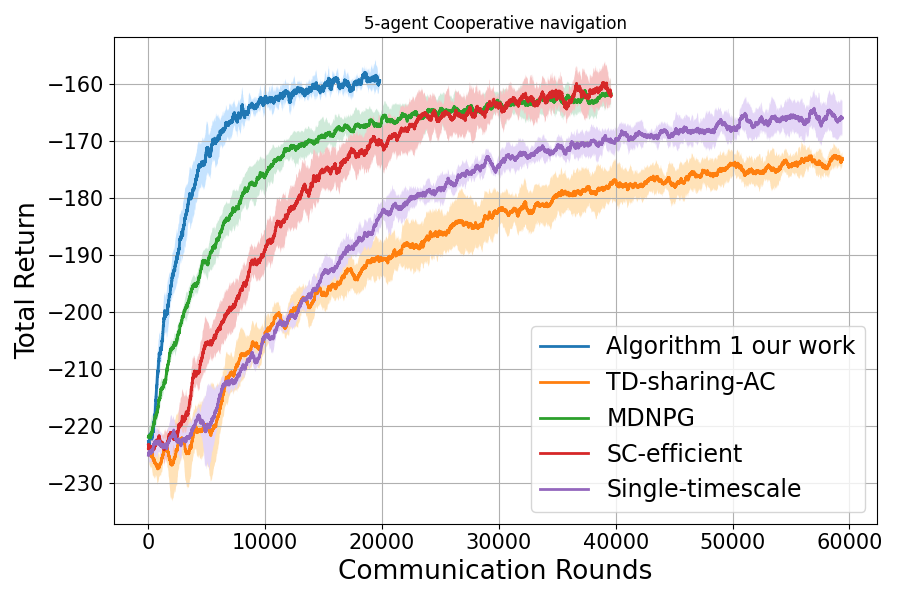}
    \caption{Comparison with related algorithms. The x-axis indicates the number of communication rounds.}
    \label{fig:comparison}
\end{figure}

\begin{figure}
  \centering
  \subfloat[Consensus Error]{\includegraphics[width=0.85\linewidth]{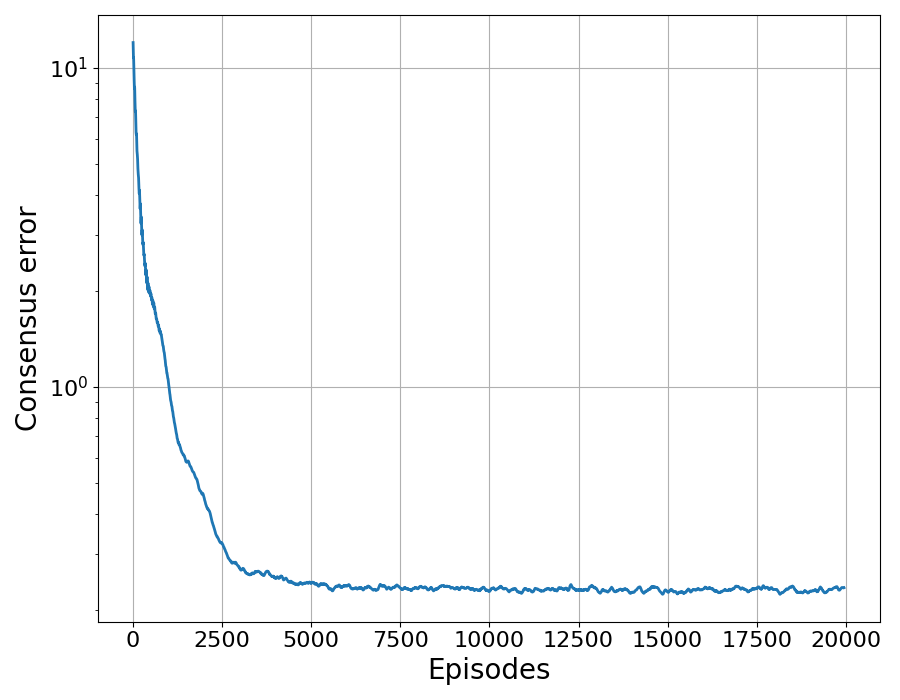}\label{fig:consensus_error}}
  \hfill
  \subfloat[Critic Loss]{\includegraphics[width=0.85\linewidth]{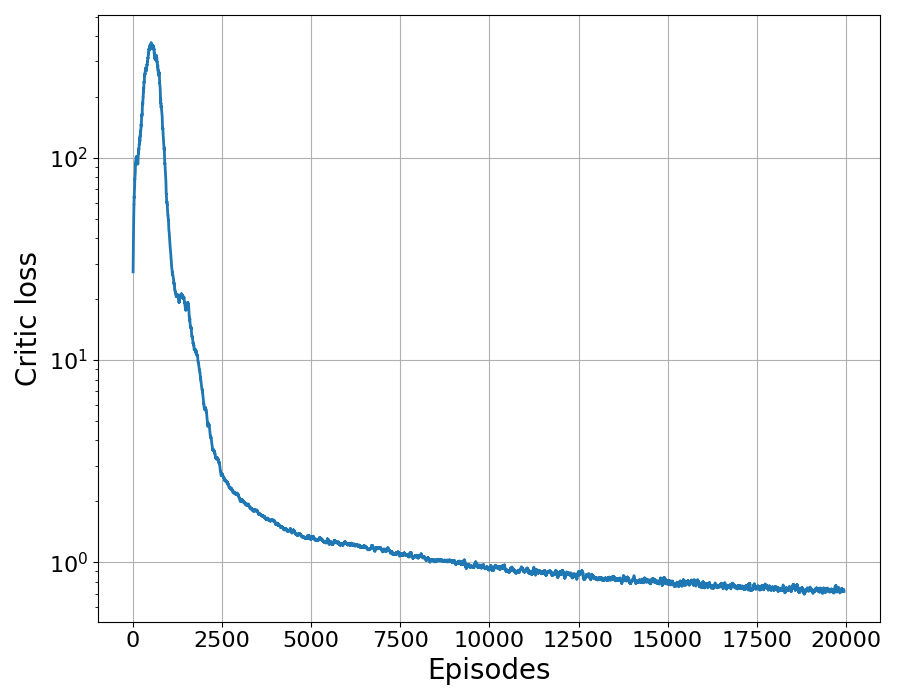}\label{fig:critic_loss}}
  \caption{Plots of the consensus error and critic loss of Algorithm~\ref{alg: LT-ADMM Actor-Critic}.}
  \label{fig:errors}
\end{figure}

Figure~\ref{fig:comparison} depicts the comparison of the algorithms in terms of the smoothed accumulated return of $20$ steps, averaged over five random seeds, and the shaded area indicates the standard deviation. 
It is shown from Figure~\ref{fig:comparison} that Algorithm~\ref{alg: LT-ADMM Actor-Critic} outperforms the other four test methods. The plots of the consensus error and critic loss of Algorithm~\ref{alg: LT-ADMM Actor-Critic} are shown in Figure~\ref{fig:errors}. It can be observed that the consensus error converges to a small bound, which is consistent with our theoretical analysis (see \eqref{sum_d}), and the critic loss also converges to a bounded approximation error.

\begin{figure}
    \centering
    \includegraphics[scale=0.4]{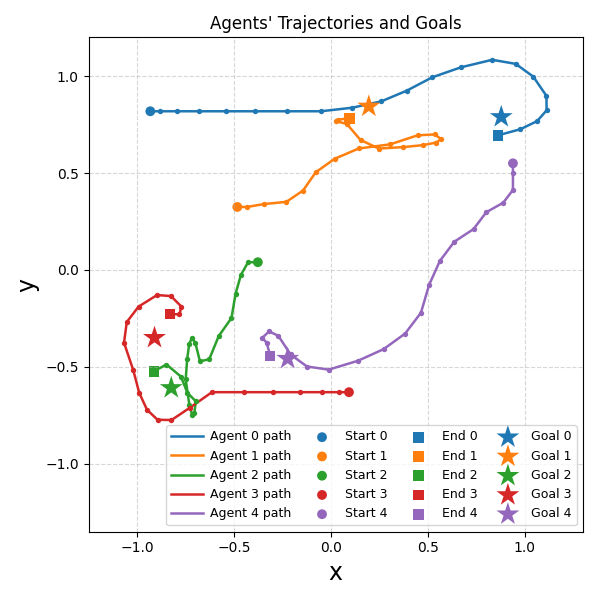}
    \caption{Trajectories of five agents using the policy learned by Algorithm~\ref{alg: LT-ADMM Actor-Critic}, case 1.}
    \label{fig:nav}
\end{figure}

\begin{figure}
    \centering
    \includegraphics[scale=0.4]{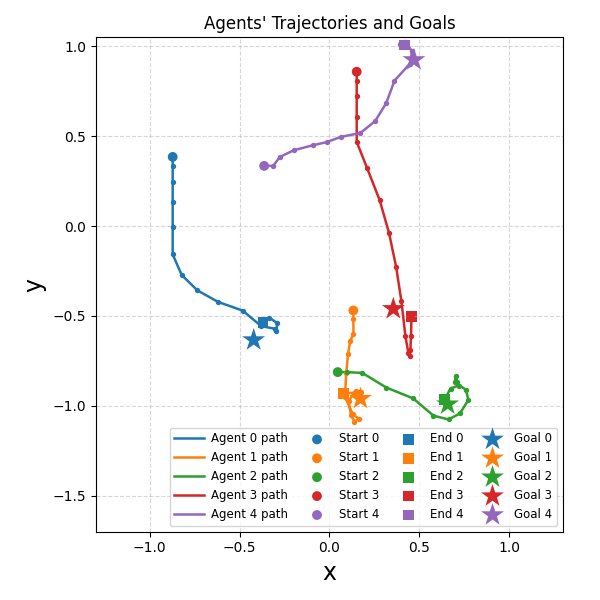}
    \caption{Trajectories of five agents using the policy learned by Algorithm~\ref{alg: LT-ADMM Actor-Critic}, case 2.}
    \label{fig:nav1}
\end{figure}

\begin{figure}
    \centering
    \includegraphics[scale=0.4]{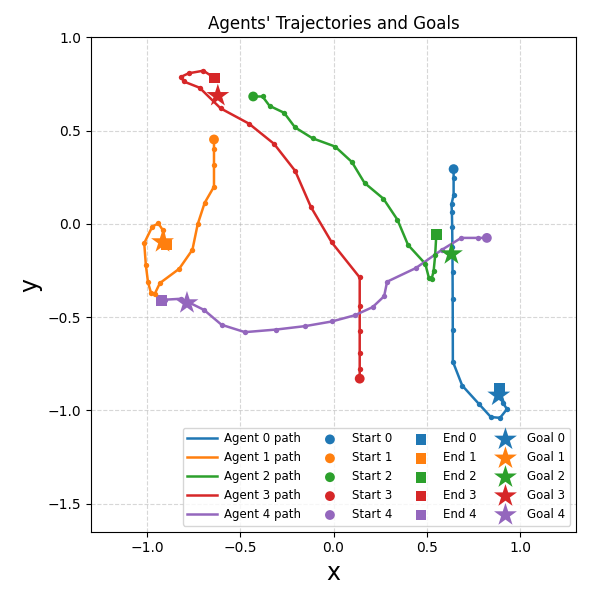}
    \caption{Trajectories of five agents using the policy learned by Algorithm~\ref{alg: LT-ADMM Actor-Critic}, case 3.}
    \label{fig:nav2}
\end{figure}

\begin{figure}
    \centering
    \includegraphics[scale=0.4]{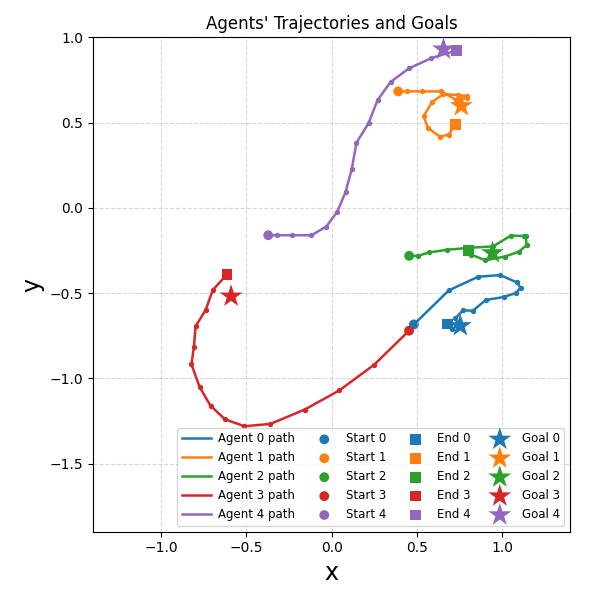}
    \caption{Trajectories of five agents using the policy learned by Algorithm~\ref{alg: LT-ADMM Actor-Critic}, case 4.}
    \label{fig:nav3}
\end{figure}

We implement the learned policy by Algorithm~\ref{alg: LT-ADMM Actor-Critic} in the navigation environment, under four different initial conditions. As shown in
Fig.~\ref{fig:nav}, Fig.~\ref{fig:nav1}, Fig.~\ref{fig:nav2} and Fig.~\ref{fig:nav3}, the agents reach their assigned landmarks without collisions within $25$ steps (sum of all agent–landmark distances is less than $0.15$).




\section{Concluding Remarks}
In this paper, we presented a communication efficient decentralized actor--critic algorithm that integrated local training and Markovian mini-batch sampling. 
In contrast to most existing decentralized actor--critic methods, the proposed framework required only the exchange of policy parameters, while critic updates were performed locally using a multi-layer neural network approximation. 
We established finite-time convergence guarantees and derived explicit bounds on the corresponding sample and communication complexities. 

Several directions for future research remain open. 
First, extending the analysis to settings with partial observability and heterogeneous information structures would allow agents to operate with local observations or belief states rather than full global states. 
Second, relaxing the assumption of symmetric and static communication networks to accommodate time-varying or directed graphs could enhance scalability and robustness in large-scale systems. 
Third, investigating asynchronous implementations of the proposed algorithm, where agents update and communicate at different rates, would make the framework more suitable for practical distributed computing environments. 
Finally, exploring extensions to continuous-time dynamics and safety-constrained reinforcement learning could help bridge the gap between theoretical guarantees and deployment in complex cyber-physical systems.

\appendices

\section{Step-size bounds}\label{sec:preliminary_definition}
The step-size upper bound for Algorithm~\ref{alg: LT-ADMM Actor-Critic} are, respectively:
\begin{equation} \label{gamma_sgd}
	\bar\alpha \coloneqq \min _{i=1, 2,  \ldots, 6} \bar\alpha_i, 
 \end{equation}
where:
\begin{align*}
\bar{\alpha}_1&\coloneqq  \min \left\lbrace  1, \frac{1}{2 L \tau \sqrt{3 }}
\right\rbrace, \ \bar{\alpha}_2 \coloneqq \frac{\sqrt{N}}{8L\tau},
 \bar{\alpha}_3\coloneqq {\frac{3}{4 L \tau}}, 
\\ \bar{\alpha}_4 &\coloneqq \frac{ \lambda_l}{\lambda_u\sqrt{16( 1+ 2\rho^2 \|\Tilde{\mathbf{L}} \|^2) \tau L^2 \|\widehat{\mathbf{V}}^{-1} \|^2 \beta_0}  }, 
\\\bar{\alpha}_5&\coloneqq \sqrt[4]{ \frac{ \lambda_l^2  }{ 16 c_4  \lambda_u^2 \tilde{c}_2 } }, \bar{\alpha}_6  \coloneqq  \sqrt[4]{\frac{\lambda_l^2\beta^2 }{48c_4 \tau \lambda_u^2  L^2 N\|\widehat{\mathbf{V}}^{-1} \|^2 } }.
\end{align*}

These bounds depend on the following constants,

\begin{align*}
 \beta_0 & \coloneqq \frac{72\beta \tau^2}{\lambda_l\rho}  + 216\tau^3\beta^2 
\\\tilde{c}_2&  \coloneqq  48 ( 1+ 2\rho^2 \|\Tilde{\mathbf{L}} \|^2) L^2 \tau^4 N \|\widehat{\mathbf{V}}^{-1} \|^2 ,
\\c_4 &\coloneqq  \frac{ 8 L^2}{ N}  \left( \frac{72\beta \tau}{\lambda_l\rho}  + 216 \tau^2\beta^2  \right) 
\end{align*}
where $ \frac{1}{\tau\lambda_u\rho} \leq \beta <  \frac{2}{\tau\lambda_u\rho}$, $\lambda_u$ and $\lambda_l$ are the largest eigenvalue and the smallest nonzero eigenvalue of the graph Laplacian matrix, respectively.

\section{Bounded policy gradient}
\begin{lemma}\label{lemma:bound_V}
\label{lem:value-bounded}
Under Assumption~\ref{ass:r_bounded}, the value function is uniformly bounded. 
Specifically, there exists a constant $C_V > 0$ such that 
\[
|V(s, \hat{\theta}_{\pi})| \le C_V, 
\quad \forall\, s \in \boldsymbol{S}, \ \hat{\theta}_{\pi} \in \Theta.
\]
\end{lemma}

\begin{proof}
    Assumption~\ref{ass:r_bounded} implies that  $\forall s \in \boldsymbol{S}$, $| \mathcal{V}_{\pi}(s) |  \leq \frac{R_{\max}}{1-\gamma}$, therefore $ | V(s, \hat\theta_\pi) | \leq \frac{R_{\max}}{1-\gamma} + \varsigma_{\mathrm{approx}}$ according to \eqref{eq:value_approximation_accuracy}, we finish the proof.
\end{proof}

The following lemma provides the bound for the critic value approximator.

\begin{lemma} \label{lem: value_function_approximation}
\label{lem:td-ms-bound}
Suppose Algorithm~\ref{alg:decentralized_td} is executed under 
Assumptions~\ref{ass:r_bounded}--\ref{ass:initial_bounded}. 
Then, for all $s \in \boldsymbol{S}$, the following bound holds:
\begin{equation}
\begin{aligned}
&(1 - \gamma)\mu_{l} \, \mathbb{E}\!\left[ 
\| V(s, \theta_{k,t}) - V(s, \hat{\theta}_{\pi_{\phi^i_{k,t}}}) \|^2 
\right]\\
&\le \frac{\| \theta_{k,t,0} - \hat{\theta}_{\pi_{\phi^i_{k,t}}} \|^2}{2 \eta T_c}
+ \mathcal{O}\!\left( \varsigma_{\mathrm{approx}} + \frac{1}{\sqrt{m}} \right)
\\
&\quad + \mathcal{O}(\eta \varsigma_{\mathrm{approx}}^2)
+ \frac{1}{(1 - \gamma)^2} \mathcal{O}(\eta)
\\
&\quad + \mathcal{O}\!\left( 
\eta \frac{\log \!\frac{\kappa}{\eta}}{1 - \zeta} 
\, \varsigma_{\mathrm{approx}}^2 \right)
+ \frac{1}{(1 - \gamma)^2} 
\mathcal{O}\!\left(
\eta \frac{\log \!\frac{\kappa}{\eta}}{1 - \zeta}
\right).
\end{aligned}
\end{equation}
\end{lemma}
\begin{proof}
Under geometric ergodicity (Assumption \ref{ass:ergodicity}), consider a Markov chain with stationary distribution $\mu_{\omega}$ induced by policy $\pi_{\omega}$. With batch size $N_c$, define
\begin{equation} \label{eq:mini_gradient}
h_{\mathrm{mb}}(\theta_t) =  \frac{1}{N_c} \sum_{i=1}^{N_c} h_i(\theta_t)
\end{equation}
where $ h_i(\theta_t) = \nabla_\theta V(s_i, \theta_t) \left( r_i + \gamma V(s_{i+1}, {\theta}_t) - V(s_i, \theta_t) \right) $, the states $\{s_i\}_{i=1}^{N_c}$ are sampled sequentially from the Markov chain with the initial distribution $\mu_{\omega}$. Then
  \begin{align}
\left\| h_{\mathrm{mb}}(\theta_t)  \right\|^2
&= \| \frac{1}{N_c} \sum_{i=1}^{N_c} h_i(\theta_t) \|^2 \le \frac{1}{N_c^2} ( \sum_{i=1}^{N_c} \|h_i(\theta_t)\| )^2 
\\& \le \frac{1}{N_c^2} \cdot N_c \sum_{i=1}^{N_c} \|h_i(\theta_t)\|^2
= \frac{1}{N_c} \sum_{i=1}^{N_c} \|h_i(\theta_t)\|^2 .
\end{align}

Taking the expectation on both sides gives
\begin{equation}
\mathbb{E}\,\| h_{\mathrm{mb}}(\theta_t) \|^2
\;\le\; \frac{1}{N_c} \sum_{i=1}^{N_c} \mathbb{E}\|h_i(\theta_t)\|^2,
\end{equation}
according to \cite[Lemma A.12]{tian2023performance} we have:
\begin{align*}
\E \norm{h_i(\theta_t)}^2 \leq \cO(\varsigma_{\mathrm{approx}}^2) + \frac{1}{(1-\gamma)^2} \cO(1),
\end{align*}
then 
\begin{equation}
\E \norm{ h_{\mathrm{mb}}(\theta_t) }^2 \leq \cO(\varsigma_{\mathrm{approx}}^2) + \frac{1}{(1-\gamma)^2} \cO(1).
\end{equation}

Then the result can be derived by replacing the gradient used in the proof of \cite[Theorem 3.1]{tian2023performance} with the mini-batch gradient $ h_{\mb}$ defined in \eqref{eq:mini_gradient}.
\end{proof}

Now we provide the variance of the policy gradient estimator in the actor update.
\begin{lemma} \label{lemma:bounded_variace}
Suppose Assumptions~\ref{ass:r_bounded}--\ref{assu:smoothness} hold. 
Then, the policy gradient estimators possess uniformly bounded variance; specifically,
for all $i \in \mathscr{V}$ and all iterations $(k,t)$,
\begin{equation}
\label{eq:pg-var-bound}
\mathbb{E}\!\left[
\big\| g_i(\phi_{k,t}^i) - \nabla J_i(\phi_{k,t}^i) \big\|^2
\right]
\le \sigma,
\end{equation}
where $\sigma$ is defined as follows,
\begin{align} \label{eq:sigma}
& \sigma= 3{C_\psi^2 (1 + \gamma)^2} \bigg(    \frac{  \epsilon_{\theta}^2}{2 \eta T_c (1 - \gamma)\mu_{l}}
+ \mathcal{O} ( \varsigma_{\mathrm{approx}} + \frac{1}{\sqrt{m}} )\nonumber
\\&+ \mathcal{O}(\eta \varsigma_{\mathrm{approx}}^2) 
+ \frac{\mathcal{O}(\eta)}{(1 - \gamma)^3} 
+ \mathcal{O} ( \eta \frac{\log \frac{\kappa}{\eta}}{1 - \rho} \varsigma_{\mathrm{approx}}^2 )
+ \frac{ \mathcal{O} ( \eta \frac{\log \frac{\kappa}{\eta}}{1 - \zeta} )}{(1 - \gamma)^3}  \bigg) \nonumber
\\& + \frac{27 C_\psi^2 (R_{\max} + 2 C_V)^2 (\kappa + 1 - \zeta)}{B(1 - \zeta)}+ 12 C_\psi^2 \varsigma_{\mathrm{approx}}^2,
\end{align}
where  $\epsilon_{\theta} = 2\tilde{\theta}$, it can be seen that mini-batch sampling can reduce the variance with $B$.
\end{lemma}
\begin{proof}
We have the following relation
\begin{equation}
\begin{aligned}
& \| g_i(\phi_{k,t}^i) - \nabla J_i(\phi_{k,t}^i) \|^2 
\leq 
 3 \| g_i(\phi_{k,t}^i)  - g_{k,t}^{*,i} \|^2 
\\&+ 3 \| g_{k,t}^{*,i} - \tilde{g}_{k,t}^{*,i} \|^2 
+ 3 \| \tilde{g}_{k,t}^{*,i} - \nabla J(\phi_{k,t}^i)\|^2,
\end{aligned}
\end{equation}
where 
\begin{equation}
\begin{aligned}
& g_{k,t}^{*,i} = \frac{1}{B} \sum_{l=tB}^{(t+1)B-1}
[ {r}_{k,l}^{i} + \gamma V(s_{k,l+1}, \hat{\theta}_{\pi_{\phi^i_{k,t}}} )  -  V(s_{k,l}, \hat{\theta}_{\pi_{\phi^i_{k,t}}} )  ] 
\\& \psi^i_{k,t}(a_{k,l} | s_{k,l}),
\\&\tilde{g}_{k,t}^{*,i}=\mathbb{E}_{s \sim \mu_{\phi^i_{k,t}},\, a \sim \pi_{\phi^i_{k,t}}(\cdot|s),\, s' \sim \mathcal{P}(\cdot|s,a)}
\\&[ {r}^i(s,a,s') + \gamma  V(s', \hat{\theta}_{\pi_{\phi^i_{k,t}}} )  - V(s, \hat{\theta}_{\pi_{\phi^i_{k,t}}})] \psi^i_{k,t}(a | s),
\end{aligned}
\end{equation}
and $\mu_{\phi^i_{k,t}}$ is the induced stationary state distribution by policy $\pi_{\phi^i_{k,t}}$.
Recall that
\begin{equation}
\begin{aligned}
g_i(\phi_{k,t}^i) &= \frac{1}{B} \sum_{l=tB}^{(t+1)B-1}
\left[ {r}_{k,l}^{i} + \gamma  V(s_{k,l+1},  \theta_{k,t}^{i} )  
-  V(s_{k,l},  \theta_{k,t}^{i} ) 
\right] \\& \quad \psi_{k,t}^{i}(a_{k,l} | s_{k,l}).
\end{aligned}
\end{equation}
The first term is bounded by
\begin{equation}
\begin{aligned}
&\| g_i(\phi_{k,t}^i) - g_{k,t}^{*,i} \|^2 
\\&= \|  \frac{1}{B} \sum_{l=tB}^{(t+1)B-1} 
( \gamma   (V(s_{k,l+1},  \theta_{k,t}^{i} )- V(s_{k,l+1}, \hat{\theta}_{\pi_{\phi^i_{k,t}}}) ) 
\\&- (V(s_{k,l},  \theta_{k,t}^{i} )- V(s_{k,l}, \hat{\theta}_{\pi_{\phi^i_{k,t}}}) )
)
\psi_{k,t}^i(a_{k,t} | s_{k,t}) \|^2  \\
&\leq \frac{C_\psi^2 (1 + \gamma)^2}{B} 
\sum_{l=tB}^{(t+1)B-1} 
\| V(s, \theta_{k,t}^{i} ) - V(s, \hat{\theta}_{\pi_{\phi^i_{k,t}}}) \|^2.  
\end{aligned}
\end{equation}

To bound the second term, using Assumptions~\ref{ass:r_bounded}, \ref{assu:smoothness} and Lemma~\ref{lemma:bound_V}, we apply \cite[Lemma D.2]{chen2022sample} with $ \| [ {r}(s,a,\tilde{s})^{i} + \gamma V(\tilde{s}, \hat{\theta}_{\pi_{\phi^i_{k,t}}})  -  V(s, \hat{\theta}_{\pi_{\phi^i_{k,t}}})  ]  \psi^i_{k,t}(a | s) \| 
\leq C_\psi (R_{\max} + 2C_V)  $ to obtain that
\begin{align}
&\mathbb{E} [  \| g_{k,t}^{*,i} - \tilde{g}_{k,t}^{*,i} \|^2 \mid \mathcal{F}_t ]
\nonumber \\ 
&\leq \frac{9 C_\psi^2 (R_{\max} + 2C_V)^2 (\kappa + 1 - \zeta)}{B(1 - \zeta)}.
\end{align}


Next, we bound the last term, using \eqref{eq:value_approximation_accuracy}, it follows that
\begin{equation}
\begin{aligned}
&\left\| \tilde{g}_{k,t}^{*,i} - \nabla J(\phi_{k,t}^i) \right\|^2 
\\&= \| \mathbb{E}_{\phi^i_{k,t}} [
( \gamma ( V(\tilde{s}, \hat{\theta}_{\pi_{\phi^i_{k,t}}}) - \mathcal{V}_{\phi^i_{k,t}}(\tilde{s})  )  \\& -  (V(s, \hat{\theta}_{\pi_{\phi^i_{k,t}}}) - \mathcal{V}_{\phi^i_{k,t}}(s) )  ) \psi^i_{k,t}(a|s)
] \|^2 \\& \leq 4 C_\psi^2 \cdot \varsigma_{\mathrm{approx}}^2.
\end{aligned}
\end{equation}
Combining the above relations yields that
\begin{align*}
&\mathbb{E} \left[ \left\| g_i(\phi_{k,t}^i) - \nabla J_i(\phi_{k,t}^i) \right\|^2 \right]
\\&\leq
\frac{27 C_\psi^2 (R_{\max} + 2 C_V)^2 (\kappa + 1 - \zeta)}{B(1 - \zeta)}
\\&+ \frac{3C_\psi^2 (1 + \gamma)^2}{B} 
\sum_{l=tB}^{(t+1)B-1} \| V(s, \theta_{k,t}^{i} ) - V(s, \hat{\theta}_{\pi_{\phi^i_{k,t}}})\|^2
\\&+ 12 C_\psi^2\varsigma_{\mathrm{approx}}^2.
\end{align*}
Substituting Lemma \ref{lem: value_function_approximation} into the above inequality, we conclude the proof.
\end{proof}


In the following section, we provide the finite-time convergence analysis of Algorithm~\ref{alg: LT-ADMM Actor-Critic}. 

\section{Preliminary transformation}
We first rewrite the algorithm in a compact form. Denote $\mathbf{Z} = \operatorname{col}\{z_{ij}\}_{i,j \in \mathcal{E}} 
    $, $ \mathbf{\Phi}_k^t = \operatorname{col}\{\phi^1_{k,t}, \phi^2_{k,t}, \ldots, \phi^N_{k,t}\}
   $, $ G(\Phi_k^t) = \operatorname{col}\{ g_1(\phi^1_{k,t}), g_2(\phi^2_{k,t}),\ldots, g_N(\phi^N_{k,t})\}
    $, $ \mathrm{\Omega}_k = \operatorname{col}\{ \omega^1,  \omega^2,\ldots,\omega^N\}
    $.
Define 
$\mathbf{A}= \operatorname{blk\,diag}\{ \mathbf{1}_{n_i} \}_{i \in \mathscr{V}} \otimes \mathbf{I}_n \in \mathbb{R}^{Mn \times Nn},$
where $n_i = |\mathcal{N}_i|$ is the degree of node $i$, and $M = \sum_i |\mathcal{N}_i|$.
$\mathbf{P} \in \mathbb{R}^{Md \times Md}$ is a permutation matrix that swaps $e_{ij}$  with $e_{ji}$. If there is an edge between nodes $i$, $j$, then $A^T[i,:]PA[:,j] = 1$, otherwise $A^T[i,:]PA[:,j] = 0$.
 Therefore $ \mathbf{A}^T\mathbf{P}\mathbf{A} = \Tilde{\mathbf{A}}$ is the adjacency matrix. 
We write Algorithm~\ref{alg: LT-ADMM Actor-Critic} in a compact form
\begin{subequations}\label{eq:compact-admm}
\begin{equation}
\mathrm{\Omega}_{k+1} = \mathrm{\Omega}_{k} + \sum_{t=0}^{\tau -1}( \alpha  G(\mathbf{\Phi}_k^t) - {\beta}(\rho \mathbf{D}\mathrm{\Omega}_k - \mathbf{A}^T \mathbf{Z}_k ) ) \label{eq:compact-admm-x}
\end{equation}
\begin{equation}
\mathbf{Z}_{k+1} =  \frac{1}{2}\mathbf{Z}_{k} - \frac{1}{2} \mathbf{P}\mathbf{Z}_k+ \rho \mathbf{P}\mathbf{A}\mathrm{\Omega}_{k+1}. \label{eq:compact-admm-z}
\end{equation}
\end{subequations}
Moreover, we introduce the following auxiliary variables
\begin{equation} \label{y_tilde_y}
\begin{aligned}
&\mathbf{Y}_k= \mathbf{A}^T \mathbf{Z}_{k} + \frac{\alpha}{\beta} \nabla \mathrm{J}(\mathrm{\bar{\Omega}}_k) -\rho \mathbf{D} \mathrm{\Omega}_k
\\&
\Tilde{\mathbf{Y}}_k= \mathbf{A}^T \mathbf{P}\mathbf{Z}_{k} -  \frac{\alpha}{\beta} \nabla \mathrm{J}(\bar{\mathrm{\Omega}}_k) - \rho \mathbf{D} \mathrm{\Omega}_k,
\end{aligned}
\end{equation}
where $\bar{\mathrm{\Omega}}_k = \mathbf{1}_N \otimes \bar{\omega}_k$, with $\bar{\omega}_k = \frac{1}{N} \mathbf{1}^T \mathrm{\Omega}_k$, $\mathbf{A}$ is   and $\mathbf{D} = \mathbf{A}^T\mathbf{A} = \operatorname{diag}\{ d_i \mathbf{I}_n \}_{i \in \mathscr{V}}$ is the degree matrix.

Multiplying both sides of \eqref{eq:compact-admm-z} by $\mathbf{1}^T$, and using the initial condition, we obtain $\mathbf{1}^T\mathbf{A}^T\mathbf{Z}_{k+1} = \rho \mathbf{1}^T \mathbf{D} \mathrm{\Omega}_{k+1}$.
As a consequence $\bar{\mathbf{Y}}_k =  \frac{\alpha}{\beta} \mathbf{1} \otimes \frac{1}{N}  \mathbf{1}^T\nabla \mathrm{J}(\bar{\mathrm{\Omega}}_k) =  \frac{\alpha}{\beta} \mathbf{1} \otimes \frac{1}{N} \sum_{i}\nabla J_i(\bar{\omega}_k)$, and \eqref{eq:compact-admm} can be further rewritten as
\begin{equation}\label{eq:compact-admm-2}
\begin{aligned}
   & \begin{bmatrix}
     \mathrm{\Omega}_{k+1}\\
    \mathbf{Y}_{k+1}\\
   \Tilde{\mathbf{Y}}_{k+1}
\end{bmatrix} = \begin{bmatrix}
    \mathbf{I}  &  \beta \tau \mathbf{I} & \mathbf{0}  \\
     \rho \Tilde{\mathbf{L}}  &  \rho \Tilde{\mathbf{L}}\beta \tau +  \frac{1}{2} \mathbf{I}  & - \frac{1}{2}\mathbf{I}  \\
   \mathbf{0}  &   - \frac{1}{2}\mathbf{I} &  \frac{1}{2} \mathbf{I}
\end{bmatrix} \otimes \mathbf{I}_n \begin{bmatrix}
     \mathrm{\Omega}_{k}\\
    \mathbf{Y}_{k}\\
   \Tilde{\mathbf{Y}}_{k}
\end{bmatrix} 
+ \mathbf{h}_k, 
\end{aligned}
\end{equation}
where 
\begin{equation} \label{eq: matrix_L}
\Tilde{\mathbf{L}} = \tilde{\mathbf{A}}- \mathbf{D}
\end{equation}
and $$
\begin{aligned}
\mathbf{h}_k &= 
\large[ \alpha\sum_{t=0}^{\tau -1}( \nabla G({\Phi}_k^t) -  \nabla \mathrm{J}(\bar{\mathrm{\Omega}}_k) ) ; \\&
\alpha  \rho  \Tilde{\mathbf{L}}\sum_{t=0}^{\tau -1}( \nabla G({\Phi}_k^t) -  \nabla \mathrm{J}(\bar{\mathrm{\Omega}}_k)   ) + \frac{\alpha}{\beta} ( \nabla \mathrm{J}(\bar{\mathrm{\Omega}}_{k+1}) -  \nabla \mathrm{J}(\bar{\mathrm{\Omega}}_{k}) ) ;\\& \frac{\alpha}{\beta} ( -\nabla \mathrm{J}(\bar{\mathrm{\Omega}}_{k+1}) +  \nabla \mathrm{J}(\bar{\mathrm{\Omega}}_{k}) ) \large].
\end{aligned}$$

\section{Deviation from the average} \label{sec:devitaion_aver}
We first introduce the following lemma regarding the smoothness of the reward function.
\begin{lemma}\cite[Proposition 5.2]{xu2020improved}  \label{lem:Jsmooth}
Under Assumption~\ref{assu:smoothness}, 
there exists a constant $L > 0$ such that the accumulated reward function $J(\omega)$ is $L$-smooth.
\end{lemma} 
\begin{lemma} \label{lem:devitaion_aver}
 Let Assumption~\ref{as:graph} hold, when  $\beta <  \frac{2}{\tau\lambda_u\rho}$,   
 \begin{equation} \label{X_Y_d} 
\|  \bar{\mathrm{\Omega}}_k- \mathrm{\Omega}_k \|^2 \leq \frac{18\beta \tau}{\lambda_l\rho} \|  \widehat{\mathbf{d}}_k\|^2, \quad \|  \bar{\mathbf{Y}}_k- \mathbf{Y}_k \|^2 \leq 9 \|  \widehat{\mathbf{d}}_k \|^2,
\end{equation}
and 
\begin{equation} \label{d_k_0}
\|\widehat{\mathbf{d}}_{k+1}\|^2  \leq \delta \|\widehat{\mathbf{d}}_{k}\|^2 +
\frac{1}{1-\delta} \|\mathbf{\widehat{h}}_{k}\|^2
\end{equation}
where $\delta = 1 - {\lambda_l\rho \tau \beta}/{2} <1$, $
\widehat{\mathbf{d}}_k = \widehat{\mathbf{V}}^{-1}
\begin{bmatrix}
\widehat{\mathbf{Q}}^T  \mathrm{\Omega}_{k};
   \widehat{\mathbf{Q}}^T \mathbf{Y}_{k};
   \widehat{\mathbf{Q}}^T \Tilde{\mathbf{Y}}_{k}
\end{bmatrix}
$ is defined in the proof.
\end{lemma}
\begin{proof}
Refer to \cite[Lemma 1]{ren2025communication}, we know that if a matrix $\widehat{\mathbf{Q}} \in \mathbf{R}^{N \times (N-1)}$ the matrix satisfying $\widehat{\mathbf{Q}} \widehat{\mathbf{Q}} ^T=\mathbf{I}_N-\frac{1}{N} \mathbf{1 1}{ }^T$,  $\widehat{\mathbf{Q}} ^T \widehat{\mathbf{Q}} =\mathbf{I}_{N-1}$  and $\mathbf{1}^T \widehat{\mathbf{Q}} =0$, $\widehat{\mathbf{Q}} ^T \mathbf{1}=0$.
We diagonalize $\mathbf{D}_i=\mathbf{V}_i \mathbf{\Delta}_i \mathbf{V}_i^{-1}$, where $\mathbf{\Delta}_i$ is the diagonal matrix of $\mathbf{D}_i$'s eigenvalues, and  
\begin{equation}
 \mathbf{V}_i= 
 \begin{bmatrix}
   -\beta \tau& d_{12}& d_{13} \\
  1&   d_{22}& d_{23}  \\
   1 &1 & 1
\end{bmatrix}
\end{equation} with
$d_{12}= -\beta\tau  + ((\beta\tilde{\lambda}_i\rho\tau(\beta\tilde{\lambda}_i\rho\tau + 2))^{0.5})/(\tilde{\lambda}_i\rho)$, $d_{13} = -\beta\tau  - ((\beta\tilde{\lambda}_i\rho\tau(\beta\tilde{\lambda}_i\rho\tau + 2))^{0.5} )/(\tilde{\lambda}_i\rho)$, $d_{22} =\tilde{\lambda}_i\rho d_{12} -1$, $d_{23}= \tilde{\lambda}_i\rho d_{13} -1$,  $\tilde{\lambda}_i<0$, $i = 1,\ldots, N-1$ is the nonzero eigenvalue of $\Tilde{\mathbf{L}}$, note that $|\tilde{\lambda}_{\min}(\Tilde{\mathbf{L}})|$ and $|\tilde{\lambda}_{\max}(\Tilde{\mathbf{L}})|$ are the largest eigenvalue and the smallest nonzero eigenvalue of the Laplacian matrix, respectively. In the following, we denote $ \lambda_l = |\tilde{\lambda}_{\max}(\Tilde{\mathbf{L}})|$ and $\lambda_u = |\tilde{\lambda}_{\min}(\Tilde{\mathbf{L}})|$.
We conclude that we can write $\mathbf{\Lambda}=(\mathbf{P}_0 \boldsymbol{\phi})^T \mathbf{V} \boldsymbol{\Delta} \mathbf{V}^{-1} (\mathbf{P}_0 \boldsymbol{\phi})$ where $\mathbf{V}=\operatorname{blkdiag}\left\{V_i\right\}_{i=2}^N $ and 
$\boldsymbol{\Delta}=\operatorname{blkdiag}\left\{\mathbf{\Delta}_i\right\}_{i=2}^N$.
Moreover, $\|\mathbf{\Delta}\| = 1 - {\lambda_l\rho \tau \beta}/{2}$
when $\lambda_u\rho \tau \beta <2$.
%
$\widehat{\mathbf{V}}^{-1}=\mathbf{V}^{-1}(\mathbf{P}_0 \boldsymbol{\phi})$, where $\mathbf{P}_0$ is a permutation matrix and $\boldsymbol{\phi}$ is an orthogonal matrix, yields
\begin{equation}\label{eq:delta-hat}
\widehat{\mathbf{d}}_{k+1}=\mathbf{\Delta} \widehat{\mathbf{d}}_k+ \widehat{\mathbf{h}}_{k},
\end{equation}
where $
\widehat{\mathbf{d}}_k = \widehat{\mathbf{V}}^{-1}
\begin{bmatrix}
\widehat{\mathbf{Q}}^T  \mathrm{\Omega}_{k};
\widehat{\mathbf{Q}}^T \mathbf{Y}_{k};
\widehat{\mathbf{Q}}^T \Tilde{\mathbf{Y}}_{k}
\end{bmatrix}$,
$\widehat{\mathbf{h}}_{k} =  \widehat{\mathbf{V}}^{-1} \widehat{\mathbf{Q}}^T  \mathbf{h}_k $. 
\end{proof}

\begin{lemma} \label{lem:phi_k}
Suppose Assumptions~\ref{as:graph}-~\ref{assu:smoothness} hold. 
If $\beta < \frac{2}{\tau \lambda_u \rho}$ and $\alpha \le \bar{\alpha}_1$, 
define
\[
\|\widehat{\boldsymbol{\Phi}}_k\|^2 
\coloneqq 
\sum_{i=1}^{N}\sum_{t=0}^{\tau-1} 
\|\phi_{k,t}^i - \bar{\omega}_k\|^2
= 
\sum_{t=0}^{\tau-1}
\|\boldsymbol{\Phi}_k^t - \bar{\boldsymbol{\Omega}}_k\|^2.
\]
Then, the following bound holds:
\begin{equation}
\label{phi_sgd}
\begin{aligned}
\mathbb{E}\!\left[\|\widehat{\boldsymbol{\Phi}}_k\|^2\right]
&\le 
\left( 
\frac{72 \beta \tau^2}{\lambda_l \rho} 
+ 216 \tau^3 \beta^2
\right)
\mathbb{E}\!\left[\|\widehat{\boldsymbol{d}}_k\|^2\right]
\\
&\quad 
+ 12 \tau^2 \alpha^2 N \sigma
+ 24 \tau^3 N \alpha^2 
\mathbb{E}\!\left[\|\nabla J(\bar{\omega}_k)\|^2\right].
\end{aligned}
\end{equation}
\end{lemma}

\smallskip
\begin{proof}
From \eqref{eq:compact-admm} we can derive that 
\begin{equation} \label{bar_x}
\begin{aligned}
\bar{\omega}_{k+1}  &= \bar{\omega}_{k}  + \frac{\alpha}{N} \sum_{t=0}^{\tau -1} \sum_{i=1}^{N} g_i(\phi_{i,k}^{t}) 
\end{aligned}
\end{equation}
and
\begin{equation}  \label{Phi_k}
      \Phi_k^{t+1} =\Phi_k^{t} + \beta \mathbf{Y}_{k} + \alpha( G(\Phi_k^t) -  \nabla \mathrm{J}(\bar{\mathrm{\Omega}}_k)   ).
\end{equation}
Recall that by Lemma \ref{lemma:bounded_variace},  $\| G(\Phi_k^t) - \nabla J(\Phi_k^t)  \|^2 \leq N\sigma$.
Now, suppose that $\tau \geq 2$, using Lemma~\ref{lem:Jsmooth} and Jensen's inequality, we obtain
\begin{align}
&  \mathbb{E} [ \left\|\Phi_k^{t+1}-\bar{\mathrm{\Omega}}_k\right\|^2 ] \nonumber 
\\& =  \mathbb{E} [ \|\Phi_k^{t}-\bar{\mathrm{\Omega}}_k  + \beta \mathbf{Y}_{k} \nonumber 
\\& + \alpha( \nabla \mathrm{J}(\mathrm{\Phi}_k^t) -  \nabla \mathrm{J}(\bar{\mathrm{\Omega}}_k)  +  G(\Phi_k^t) - \nabla \mathrm{J}(\mathrm{\Phi}_k^t)) \|^2 ] \nonumber 
\\& \leq\left(1+\frac{1}{\tau-1} \right)  \mathbb{E} [ \left\|\Phi_k^{t}-\bar{\mathrm{\Omega}}_k\right\|^2 ] \nonumber 
\\&+ \tau   \mathbb{E} [\|  \beta \mathbf{Y}_{k} + \alpha( \nabla \mathrm{J}(\mathrm{\Phi}_k^t) -  \nabla \mathrm{J}(\bar{\mathrm{\Omega}}_k)  +  G(\Phi_k^t) - \nabla \mathrm{J}(\mathrm{\Phi}_k^t) ) \|^2 ] \nonumber 
\\& \leq \left( 1 + \frac{1}{\tau-1} + 3\alpha^2 \tau  L^2   \right)  \mathbb{E} [ \left\|\mathbf{\Phi}_{k}^t-\bar{\mathrm{\Omega}}_k\right\|^2 ] \nonumber 
\\&+3 \tau \beta^2   \mathbb{E} [ \left\|  \mathbf{Y}_{k}  \right\|^2 ]
 + 3\tau \alpha^2\mathbb{E}[\|  G(\Phi_k^t) - \nabla \mathrm{J}(\mathrm{\Phi}_k^t)  \|^2 ] \nonumber  
\\& \leq\left(1+\frac{5 / 4}{\tau-1}\right)  \mathbb{E} [\|\mathbf{\Phi}_{k}^t-\bar{\mathrm{\Omega}}_k\|^2 ] \nonumber
\\&+3 \tau \beta^2   \mathbb{E} [ \left\|  \mathbf{Y}_{k}  \right\|^2 ]
 + 3\tau \alpha^2\mathbb{E}[\|  G(\Phi_k^t) - \nabla \mathrm{J}(\mathrm{\Phi}_k^t)  \|^2 ], 
\end{align}
where the last inequality holds when 
\begin{equation} \label{eq.gamma_saga_1}
3 \alpha^2 \tau L^2 \leq \frac{1/4}{\tau -1},   
\end{equation}
which is satisfied by $\alpha\leq \bar{\alpha}_1$.

Iterating the above inequality for $t=0,\ldots, \tau-1$ 
\begin{align*}
\mathbb{E} & [  \left\|\Phi_k^{t+1}-\bar{\mathrm{\Omega}}_k\right\|^2] \leq \left(1+\frac{5 / 4}{\tau-1}\right) ^t \mathbb{E} [\left\|\mathrm{\Omega}_{k}-\bar{\mathrm{\Omega}}_k\right\|^2] +
\\& +3 \tau \beta^2 \sum_{l=0}^t  \left(1+\frac{5 / 4}{\tau-1}\right) ^l \mathbb{E} [ \left\| \mathbf{Y}_{k} -\bar{\mathbf{Y}}_k + \bar{\mathbf{Y}}_k \right\|^2 ] 
\\&
+  3 \tau \alpha^2 \sum_{l=0}^t  \left(1+\frac{5 / 4}{\tau-1}\right) ^l N \sigma
\\&\leq 4 \mathbb{E} [ \left\|\mathrm{\Omega}_{k}-\bar{\mathrm{\Omega}}_k\right\|^2 ] 
+ 12\tau  \alpha^2 N \sigma 
\\&+ 12\tau^2\beta^2 \mathbb{E} [ \left\| \mathbf{Y}_{k} -\bar{\mathbf{Y}}_k + \bar{\mathbf{Y}}_k  \right\|^2],
\end{align*}
where the last inequality holds by $(1+ \frac{a}{\tau -1})^t \leq \exp(\frac{at}{\tau -1})\leq \exp(a)$ for $t\leq \tau-1$ and $a = {5}/{4}$.

Summing over $t$, it follows that
\begin{equation}\label{phi_sgd_0}
\begin{split}
 & \mathbb{E} [ \|\widehat{\mathbf{\Phi}}_k \|^2 ]\leq 
4\tau \mathbb{E} [\|\mathrm{\Omega}_{k}-\bar{\mathrm{\Omega}}_k\|^2 ]   +  12\tau^2 \alpha^2 N \sigma  \\& + 24\tau^3\beta^2 \mathbb{E} [\| \mathbf{Y}_{k} -\bar{\mathbf{Y}}_k \|^2 ] + 24\tau^3 N\alpha^2 \mathbb{E} [  \| \nabla J(\bar{\omega}_k) \|^2 ],
\end{split}
\end{equation}
moreover, it is easy to verify that \eqref{phi_sgd_0} also holds for $\tau =1$.
Using   \eqref{X_Y_d} concludes the proof.
\end{proof}

\begin{lemma}
Suppose Assumptions~\ref{as:graph}-~ \ref{assu:smoothness} hold. If $\beta <  \frac{2}{\tau\lambda_u\rho}$ and $\alpha \leq \bar{\alpha}_1$, then
\begin{equation} \label{d_k}
\begin{aligned}
&\mathbb{E} [ \|\widehat{\mathbf{d}}_{k+1}\|^2 ] 
\\&\leq ( \delta + \frac{c_0}{1-\delta}) \mathbb{E} [ \|\widehat{\mathbf{d}}_{k}\|^2 ] 
+ \frac{c_1}{1-\delta}  \mathbb{E}[ \| \sum_t \frac{1}{N}  \sum_{i=1}^{N} g_i (\phi_{k,t}^i) \|^2 ]
\\& + \frac{c_2}{1-\delta}\mathbb{E}[\| \nabla J(\bar{\omega}_k) \|^2] + \frac{c_3}{1-\delta} \sigma
\end{aligned} 
\end{equation}
where 
\begin{align*}
\delta & \coloneqq 1- \frac{\lambda_l\rho \tau \beta}{2}, 
\\ \beta_0 & \coloneqq \frac{72\beta \tau^2}{\lambda_l\rho}  + 216\tau^3\beta^2 
\\c_0 & \coloneqq \alpha^2 ( 1+ 2\rho^2 \|\Tilde{\mathbf{L}} \|^2)  2\tau L^2 \|\widehat{\mathbf{V}}^{-1} \|^2 \beta_0, 
\\ 
c_1 & \coloneqq 3 L^2 \frac{\alpha^4}{\beta^2} N\|\widehat{\mathbf{V}}^{-1} \|^2 ,
\\c_2&  \coloneqq \alpha^2 ( 1+ 2\rho^2 \|\Tilde{\mathbf{L}} \|^2)  2\tau L^2 24\tau^3 N \alpha^2 \|\widehat{\mathbf{V}}^{-1} \|^2 ,
\\ 
c_3 & \coloneqq \alpha^2 ( 1+ 2\rho^2 \|\Tilde{\mathbf{L}} \|^2)  2\tau^2 N  \|\widehat{\mathbf{V}}^{-1} \|^2  
\\&+  \alpha^2 ( 1+ 2\rho^2 \|\Tilde{\mathbf{L}} \|^2) 2\tau L^2 12\tau^2\alpha^2 N \|\widehat{\mathbf{V}}^{-1} \|^2,
\end{align*}
\end{lemma}
\begin{proof}
According to Lemma~\ref{lem:Jsmooth}, there exists a constant $L$ such that the accumulated reward function $J(\omega)$ is $L$-smooth.
Therefore, 
\begin{align*}
&\| \sum_{t=0}^{\tau -1}( G(\mathrm{\Phi}_k^t) -  \nabla \mathrm{J}(\bar{\mathrm{\Omega}}_k)   )\|^2
\\& =   \| \sum_{t=0}^{\tau -1}( G(\mathrm{\Phi}_k^t) - \nabla \mathrm{J}(\mathrm{\Phi}_k^t)  +   \nabla \mathrm{J}(\mathrm{\Phi}_k^t) -  \nabla \mathrm{J}(\bar{\mathrm{\Omega}}_k)   )\|^2
\\&
\leq 2\tau L^2 \|\widehat{\mathbf{\Phi}}_k \|^2
 +   2\tau \sum_{t=0}^{\tau -1} \| G(\mathrm{\Phi}_k^t)  - \nabla J(\mathrm{\Phi}_k^t)   \|^2
\end{align*}
We also have
$\|\nabla \mathrm{J}(\bar{\mathrm{\Omega}}_{k+1}) - \nabla \mathrm{J}(\bar{\mathrm{\Omega}}_{k}) \|^2 \leq N L^2 \left\| \bar{\omega}_{k+1} - \bar{\omega}_{k} \right\|^2 
 = N  L^2 \alpha^2 \|\frac{1}{N} \sum_{t=1}^{\tau} \sum_{i=1}^{N} g_i (\phi_{k,t}^i) \|^2
$, it further holds that
\begin{align} \label{eq:h_k_0}
    & \|{\mathbf{h}}_{k}\|^2 
 \leq \alpha^2 ( 1+ 2\rho^2 \|\Tilde{\mathbf{L}} \|^2)  ( 2\tau L^2 \|\widehat{\mathbf{\Phi}}_k \|^2 \nonumber
\\&+ 2\tau \sum_{t=0}^{\tau -1} \| G(\mathrm{\Phi}_k^t)  - \nabla J(\mathrm{\Phi}_k^t)   \|^2 ) \nonumber
\\& +3 N L^2 \frac{\alpha^4}{\beta^2} \bigg(  \| \frac{1}{N}  \sum_t  \sum_{i=1}^{N} g_i (\phi_{k,t}^i)\|^2 \bigg)\nonumber
\\&  \leq \alpha^2 ( 1+ 2\rho^2 \|\Tilde{\mathbf{L}} \|^2)  ( 2\tau L^2 \|\widehat{\mathbf{\Phi}}_k \|^2 + 2\tau^2 N \sigma) \nonumber
\\& +\frac{ 3 N L^2 \alpha^4}{\beta^2} \| \frac{1}{N} \sum_{t=1}^{\tau}  \sum_{i=1}^{N} g_i (\phi_{k,t}^i) \|^2.
\end{align}
%
When $\beta <  \frac{2}{\tau\lambda_u\rho}$ and $\alpha \leq \bar{\alpha}_1$, using \eqref{phi_sgd} we have 
\begin{align*}
\|\widehat{\mathbf{h}}_{k}\|^2 
 &\leq  c_0  \|\widehat{\mathbf{d}}_{k}\|^2 + c_1  \|\sum_t \frac{1}{N}  \sum_{i=1}^{N} g_i (\phi_{k,t}^i) \|^2 
 \\&+ c_2\mathbb{E}[\| \nabla J(\bar{\omega}_k) \|^2] +  c_3 \sigma,
\end{align*}
together with  \eqref{d_k_0}, and by $\alpha<1$, we can then derive that \eqref{d_k} holds.
\end{proof}

\section{Main result} \label{appen:main_result}

We start our proof by recalling that the following inequality holds for $L$-smooth function $J$, $\forall y, z$ \cite{nesterov2013introductory}:
\begin{equation} \label{nonconvex_inequality}
J(y) \geq J(z) + \langle \nabla J(z), y-z \rangle - (L/2) \Vert y-z  \Vert^2. 
\end{equation}
{According to Lemma~\ref{lem:Jsmooth}, there exists a constant $L$ such that $J$ is $L$-smooth. } 
Based on \eqref{bar_x}, substituting $y=\bar{\omega}_{k+1}$ and $z=\bar{\omega}_{k}$ into \eqref{nonconvex_inequality},
\begin{align*}
  \mathbb{E}  [ J \left(\bar{\omega}_{k+1}\right)  ]
 &\geq  \mathbb{E} [ J(\bar{\omega}_{k}) ] + \alpha [ \langle\nabla J\left(\bar{\omega}_{k}\right), \frac{1}{N} \sum_t \sum_i g_i(\phi_{k,t}^i)  ) \rangle ] \\& - \frac{\alpha^2 \tau L}{2} \sum_t  \|\frac{1}{N} \sum_i g_i\left(\phi_{k,t}^i \right)\|^2 .
\end{align*}
Using now $2\langle a, b\rangle=\|a\|^2+\|b\|^2-\|a-b\|^2$, we have
\begin{align*}
& \langle\nabla J\left(\bar{\omega}_{k}\right), \frac{1}{N} \sum_t \sum_i g_i\left(\phi_{k,t}^i\right) \rangle 
\\& =\frac{\tau}{2}\left\|\nabla J\left(\bar{\omega}_{k}\right)\right\|^2 +\frac{1}{2} \sum_t\|\frac{1}{N} \sum_i g_i\left(\phi_{k,t}^i\right)\|^2
\\& - \frac{1}{2} \sum_t\|\frac{1}{N} \sum_i g_i\left(\phi_{k,t}^i\right)-\nabla J\left(\bar{\omega}_{k}\right)\|^2 
\end{align*}
Now, combining the two equations above and using \eqref{X_Y_d}, yields
\begin{align*}
 & \mathbb{E} [ J\left(\bar{\omega}_{k+1}\right) ] 
 \geq   \mathbb{E} [ J\left(\bar{\omega}_{k}\right) ] + \frac{\alpha \tau}{2} \mathbb{E} [ \left\|\nabla J\left(\bar{\omega}_{k}\right)\right\|^2 ]
 \\&  + \frac{\alpha}{2}(1 -  \alpha \tau L) \sum_t \mathbb{E} [ \|\frac{1}{N} \sum_i g_i\left(\phi_{k,t}^i\right) \|^2 ]
 \\& - \frac{\alpha}{2} \sum_t\|\frac{1}{N} \sum_i g_i\left(\phi_{k,t}^i\right)-\nabla J\left(\bar{\omega}_{k}\right)\|^2 
\\& \geq  \mathbb{E} [ J\left(\bar{\omega}_{k}\right) ]  + \frac{\alpha \tau}{2} \mathbb{E} [ \left\|\nabla J\left(\bar{\omega}_{k}\right)\right\|^2 ]
 \\& +\frac{\alpha}{2}(1 -  \alpha \tau L) \sum_t \mathbb{E} [ \|\frac{1}{N} \sum_i g_i\left(\phi_{k,t}^i\right) \|^2 ]
 \\& - \alpha \sum_t\|\frac{1}{N} \sum_i g_i\left(\phi_{k,t}^i\right)-
 \frac{1}{N} \sum_i \nabla J_i\left(\phi_{k,t}^i\right) \|^2 
 \\& - \alpha \sum_t\|\frac{1}{N} \sum_i \nabla J_i\left(\phi_{k,t}^i\right)-
 \nabla J\left(\bar{\omega}_{k}\right)\|^2 
 \\& \geq \mathbb{E} [ J\left(\bar{\omega}_{k}\right) ]+\frac{\alpha \tau}{2} \mathbb{E} [ \left\|\nabla J\left(\bar{\omega}_{k}\right)\right\|^2 ]
 \\&+ \frac{\alpha}{2}(1 -  \alpha \tau L) \sum_t \mathbb{E} [ \|\frac{1}{N} \sum_i g_i\left(\phi_{k,t}^i\right) \|^2 ]
 \\& - \alpha \tau \sigma
 - \frac{\alpha L^2}{N}  \|\widehat{\mathbf{\Phi}}_k\|^2.
\end{align*}

Substituting \eqref{phi_sgd} into the above inequality yields
\begin{align*}
&  J\left(\bar{\omega}_{k+1}\right) \geq   J\left(\bar{\omega}_{k}\right) + \\& +\frac{\alpha\tau}{2}\left( 1 - \frac{48 L^2 \tau^2 \alpha^2}{N} \right)  \|\nabla J\left(\bar{\omega}_{k}\right) \|^2 
\\& +\frac{\alpha}{2}(1 - \alpha L \tau) \sum_t \mathbb{E} [ \|\frac{1}{N} \sum_i g_i\left(\phi_{k,t}^i\right)\|^2 ]
\\& - \frac{\alpha L^2}{N}  \left( \frac{72\beta \tau^2}{\lambda_l\rho}  + 216\tau^3\beta^2  \right) \mathbb{E} [ \|\widehat{\mathbf{d}}_k \|^2 ]
\\& - \alpha \tau \sigma - 12 \tau^2 \alpha^3 L^2 \sigma.
\end{align*}
When $\alpha \leq \min \lbrace \bar{\alpha}_2, \bar{\alpha}_3 \rbrace$, then 
\begin{equation} \label{gamma_SGD_2}
\frac{48 L^2 \tau^2 \alpha^2}{N} \leq \frac{3}{4}, \quad  \alpha L \tau \leq \frac{3}{4},
\end{equation}
and we can upper bound the previous inequality by
\begin{align*}
& \mathbb{E} [ J\left(\bar{\omega}_{k+1}\right) ] \geq   \mathbb{E} [ J\left(\bar{\omega}_{k}\right) ]+\frac{\alpha \tau}{8} \mathbb{E} [ \|\nabla J\left(\bar{\omega}_{k}\right) \|^2 ] +
\\&+\frac{\alpha}{8} \sum_t \mathbb{E} [ \|\frac{1}{N} \sum_{i=1}^N g_i\left(\phi_{k,t}^i\right)\|^2 ] 
\\&- \frac{\alpha L^2}{ N}  \left( \frac{72\beta \tau^2}{\lambda_l\rho}  + 216\tau^3\beta^2  \right) \mathbb{E} [ \|\widehat{\mathbf{d}}_k \|^2 ]
\\&-\alpha \tau \sigma - 12 \tau^2 \alpha^3 L^2 \sigma.
\end{align*}
Rearranging the above relation, we get
\begin{align*}
&\mathcal{D}_k  
\leq \frac{8}{\alpha \tau}  \mathbb{E} \left[ \left( {J}\left(\bar{\omega}_{k+1}\right)-  {J}\left(\bar{\omega}_{k}\right)\right) \right]\\& 
+\frac{8}{\alpha \tau} \frac{\alpha L^2}{N}  \left( \frac{72\beta \tau^2}{\lambda_l\rho}  + 216\tau^3\beta^2  \right)  \mathbb{E} [\|\widehat{\mathbf{d}}_k \|^2 ] 
 \\& + {8  \sigma} + 96 \alpha^2 \tau L^2\sigma ,
\end{align*}
where $\mathcal{D}_k$ is defined as
\begin{equation}\label{eq:convergence-metric}
\mathcal{D}_k = \mathbb{E} [  \|\nabla J (\bar{\omega}_{k} ) \|^2+\frac{1}{\tau} \sum_{t = 0}^{\tau-1} \|{\frac{1}{N} \sum_{i = 1}^N g_i (\phi_{k,t}^i )}\|^2 ].
\end{equation}
Summing over $k=0,1, \ldots, K-1$,  we obtain
\begin{align}
& \sum_{k=0}^{K-1} \mathcal{D}_k \leq  \frac{8( {J}\left(\bar{\omega}_K\right) - {J}\left(\bar{\omega}_0\right)   )}{\alpha \tau} +  c_4 \sum_{k=0}^{K-1} \mathbb{E} [ \|\widehat{\mathbf{d}}_k\|^2 ] + Kc_5 \sigma, \label{sum_E}
\end{align}
where
\begin{equation} 
\begin{aligned}
c_4 &\coloneqq  \frac{ 8 L^2}{ N}  \left( \frac{72\beta \tau}{\lambda_l\rho}  + 216 \tau^2\beta^2  \right) 
\\ c_5 & \coloneqq 8+ 96 \alpha^2 \tau L^2.
\end{aligned}
\end{equation}

We now bound the term $\sum_{k=0}^{K-1}  \|\widehat{\mathbf{d}}_k\|^2$.
From \eqref{d_k}, we have
\begin{equation} \label{d_iter}
\begin{aligned}
&\mathbb{E} [ \|\widehat{\mathbf{d}}_{k+1}\|^2 ] \\&\leq ( \delta + \frac{c_0}{1-\delta}) \mathbb{E} [ \|\widehat{\mathbf{d}}_{k}\|^2 ] 
+ \frac{c_1}{1-\delta}  \mathbb{E}[ \| \sum_t \frac{1}{N}  \sum_{i=1}^{N} g_i (\phi_{k,t}^i)  \|^2 ]
\\& + \frac{c_2}{1-\delta}\mathbb{E}[\| \nabla J(\bar{\omega}_k) \|^2] + \frac{c_3}{1-\delta} \sigma
\\& \leq  \bar{\delta}
\mathbb{E} [ \| \widehat{\mathbf{d}}_k \|^2 ] + \frac{c_3}{1-\delta} \sigma
+ R \mathcal{D}_k  
\end{aligned} 
\end{equation}
where 
\begin{equation} \label{eq:K}
\begin{split}
R &\coloneqq \operatorname{\max}\{ \frac{c_2 }{1-\delta}, \frac{c_1\tau}{1-\delta}\}.
\end{split}
\end{equation}
Moreover, letting  $\alpha \leq  \bar{\alpha}_4$ and $ \frac{1}{\tau\lambda_u\rho} \leq \beta <  \frac{2}{\tau\lambda_u\rho}$, 
we have
\begin{equation} \label{gamma_SGD_3}
\bar{\delta} =  \delta + \frac{ c_0 }{1-\delta} <1 - \frac{\lambda_l}{4\lambda_u}.
\end{equation}
and it follows that $c_4 = \frac{ 8 L^2}{ N}  ( \frac{72\beta \tau}{\lambda_l\rho}  + 216 \tau^2\beta^2  ) \leq  \frac{ 8 L^2}{ N} (\frac{144}{\lambda_l\lambda_u\rho^2} + \frac{864}{ \lambda_u^2\rho^2}  )$.
Iterating  \eqref{d_iter} now gives
$$
\mathbb{E} [ \|\widehat{\mathbf{d}}_k\|^2 ] \leq \bar{\delta}^k \mathbb{E} [ \|\widehat{\mathbf{d}}_0\|^2]+ R \sum_{\ell=0}^{k-1}\bar{\delta}^{k-1-\ell} \mathcal{D}_{\ell} + \frac{c_3 \sigma}{(1-\bar{\delta})(1-\delta)}
$$
and summing this inequality over $k=0, \ldots, K-1$, it follows that 
\begin{equation} \label{sum_d}
\begin{aligned}
\sum_{k=0}^{K-1}  \mathbb{E} [\|\widehat{\mathbf{d}}_k\|^2 ]
\leq \frac{\|\widehat{\mathbf{d}}_0\|^2}{1-\bar{\delta}  }+\frac{R}{ 1- \bar{\delta} } \sum_{k=0}^{K-1} \mathcal{D}_k + \frac{c_3 \sigma K}{(1-\bar{\delta})(1-\delta)}.
\end{aligned}
\end{equation}
It can be seen that  when $\mathcal{D}_k $ is bounded, the consensus error $\|\widehat{\mathbf{d}}_k\|^2$ is also bounded.

Substituting \eqref{sum_d} into \eqref{sum_E} and rearranging, we obtain
$$
\begin{aligned}
&\left(1- q_0\right)  \sum_{k=0}^{K-1} \mathcal{D}_k \leq  \frac{ 8( {J}\left(\bar{\omega}_K\right) - {J}\left(\bar{\omega}_0\right)   )}{\alpha \tau} + q_1 \|\hat{\mathbf{d}}_0\|^2+ Kq_2 \sigma,
\end{aligned}
$$
where 
\begin{equation}
    \begin{aligned}
& q_0 \coloneqq \frac{ c_4 R}{1-\bar{\delta}}, \quad q_1 \coloneqq  \frac{c_4}{1-\bar{\delta}} \quad q_2 \coloneqq\frac{c_4 c_3 }{(1-\bar{\delta})(1-\delta)} + c_5.
    \end{aligned}
\end{equation}
Since
$1- \bar{\delta} \geq \frac{\lambda_l}{4\lambda_u},$  when   $\alpha \leq \min \lbrace 1, \bar{\alpha}_5, \bar{\alpha}_6 \rbrace$, we have
\begin{equation} \label{gamma_SGD_4}
q_0 \leq \frac{1}{2},
\end{equation}
$q_1 \leq \frac{ 32 L^2}{ N} (\frac{144}{\lambda_l^2\rho^2} + \frac{864}{\lambda_u\lambda_l\rho^2}  ) $,
it follows that
\begin{equation} \label{eq:converge}
\begin{aligned}
& \frac{1}{K} \sum_{k=0}^{K-1} \mathcal{D}_k \leq \frac{16 ( {J}\left(\bar{\omega}_K\right) - {J}\left(\bar{\omega}_0\right)   )}{\alpha \tau K} + \frac{2q_1}{K} \|\hat{\mathbf{d}}_0\|^2+ 2q_{2}\sigma.
\end{aligned}
\end{equation}
By collecting all step-size conditions,  if the step-size $\alpha < \bar\alpha \coloneqq \min _{i=1, 2,  \ldots, 6} \bar\alpha_i$,  then  \eqref{eq:converge}  holds, the policy parameters $\{\mathrm{\Omega}_k\}$ converge to the neighborhood of the stationary point of problem \eqref{main_problem}, concluding the proof.

\balance

\bibliographystyle{IEEEtran}
\bibliography{References/reference,References/ids,References/TAC_Ref_Andreas}

\end{document}